\documentclass[11pt]{article}

\usepackage[]{EMNLP2023}

\usepackage{times}
\usepackage{latexsym}

\usepackage[T1]{fontenc}

\usepackage[utf8]{inputenc}

\usepackage{microtype}

\usepackage{microtype}
\usepackage{graphicx}
\usepackage{subcaption}
\usepackage{wrapfig}
\usepackage{booktabs} 
\usepackage[utf8]{inputenc}
\usepackage[T1]{fontenc}    %

\renewcommand{\cite}{\citep}
\usepackage{amsmath}
\usepackage{amssymb}
\usepackage{mathtools}
\usepackage{amsthm}

\usepackage[capitalize,noabbrev]{cleveref}

\usepackage{listings}
\usepackage{comment}
\usepackage{adjustbox}
\usepackage{xspace}
\usepackage{fancyvrb}
\usepackage[subtle]{savetrees}

\usepackage{times}
\usepackage{latexsym}
\usepackage[utf8]{inputenc} 
\usepackage[T1]{fontenc}    
\usepackage{hyperref}       
\usepackage{url}            
\usepackage{booktabs}       
\usepackage{amsfonts}       
\usepackage{nicefrac}       
\usepackage{microtype}      
\usepackage{xcolor}         
\usepackage[colorinlistoftodos,textsize=tiny]{todonotes}

\usepackage{amsmath}
\newcommand{\ourPrompt}{\texttt{MVP\;}}
\newcommand{\ourPromptns}{\texttt{MVP}}

\newcommand{\linearFT}{\texttt{MLP-FT\;}}
\newcommand{\linearFTns}{\texttt{MLP-FT}}
\newcommand{\CLSPrompt}{\texttt{CLSPrompt\;}}
\newcommand{\CLSPromptns}{\texttt{CLSPrompt}}
\newcommand{\sparseCLS}{\texttt{ProjectCLS\;}}
\newcommand{\sparseCLSns}{\texttt{ProjectCLS}}
\newcommand{\sparseLPFT}{\texttt{LPFT\;}}
\newcommand{\sparseLPFTns}{\texttt{LPFT}}
\newcommand{\denseLPFT}{\texttt{DenseLPFT\;}}

\newcommand{\LPFT}{\texttt{LPFT\;}}
\newcommand{\LPFTns}{\texttt{LPFT}}
\newcommand{\mask}{\texttt{[MASK]}}
\usepackage{multirow}
\usepackage{booktabs}
\usepackage[normalem]{ulem}
\useunder{\uline}{\ul}{}
\usepackage[T1]{fontenc}

\newcommand{\tablescale}{0.85}

\newlength{\widebarargwidth}
\newlength{\widebarargheight}
\newlength{\widebarargdepth}

\ifthenelse{\isundefined{\definition}}{\newtheorem{definition}{Definition}}{}
\ifthenelse{\isundefined{\assumption}}{\newtheorem{assumption}{Assumption}}{}
\ifthenelse{\isundefined{\hypothesis}}{\newtheorem{hypothesis}{Hypothesis}}{}
\ifthenelse{\isundefined{\proposition}}{\newtheorem{proposition}{Proposition}}{}
\ifthenelse{\isundefined{\theorem}}{\newtheorem{theorem}{Theorem}}{}
\ifthenelse{\isundefined{\lemma}}{\newtheorem{lemma}{Lemma}}{}
\ifthenelse{\isundefined{\corollary}}{\newtheorem{corollary}{Corollary}}{}
\ifthenelse{\isundefined{\alg}}{\newtheorem{alg}{Algorithm}}{}
\ifthenelse{\isundefined{\example}}{\newtheorem{example}{Example}}{}

\def\arxiv{1}
\ifnum\arxiv=1

\else

\fi

\begin{document}

\title{Model-tuning Via Prompts\\ Makes NLP Models Adversarially Robust}


\author{Mrigank Raman\thanks{~~~Equal contribution.} \\
  Carnegie Mellon University\\
  \small \texttt{mrigankr@cmu.edu} \\\And
   Pratyush Maini$^*$\\
  Carnegie Mellon University \\
   \small \texttt{pratyushmaini@cmu.edu} \\\And
  Zico Kolter \\
  Carnegie Mellon University \\
   \small \texttt{zkolter@cs.cmu.edu} \\
   \AND
    Zachary C. Lipton \\
  Carnegie Mellon University \\
  \small \texttt{zlipton@cmu.edu} \\\And
  Danish Pruthi  \\
  Indian Institute of Science (IISc) \\
   \small \texttt{danishp@iisc.ac.in} \\
}

\date{}

\newcommand{\fix}{\marginpar{FIX}}
\newcommand{\new}{\marginpar{NEW}}

\maketitle
\begin{abstract}


In recent years,
NLP practitioners have converged
on the following practice:
(i) import an off-the-shelf pretrained (masked) language model;
(ii) append a multilayer perceptron atop the CLS token's hidden representation
(with randomly initialized weights);
and (iii) fine-tune the entire model on a downstream task (\linearFTns).
This 
procedure 
has 
produced massive gains 
on standard NLP benchmarks,
but these models remain brittle, even to 
mild adversarial perturbations.
In this work, we demonstrate surprising gains 
in adversarial robustness enjoyed by 
Model-tuning Via Prompts (\ourPromptns),
an alternative method of adapting to downstream tasks.
Rather than appending an MLP head to make output prediction, \ourPrompt appends a prompt template to the input, and makes prediction via text infilling/completion.
Across 5 NLP datasets, 4 adversarial attacks, and 3 different models, 
\ourPrompt improves performance against adversarial 
substitutions  by an average of $8\%$ 
over standard methods and even outperforms 
adversarial training-based state-of-art defenses by $3.5\%$.
By combining \ourPrompt with adversarial training, 
we achieve further improvements in adversarial robustness
while maintaining performance on unperturbed examples. 
Finally, we conduct ablations to investigate 
the mechanism underlying these gains.
Notably, we find that the main causes of vulnerability of \linearFT 
can be attributed to the misalignment between pre-training and fine-tuning tasks, 
and the randomly initialized MLP parameters.\footnote{Code is available at \href{https://github.com/acmi-lab/mvp}{https://github.com/acmi-lab/mvp}.}

\end{abstract}

\section{Introduction}
\begin{figure*}[t]
\centering
\includegraphics[width=0.75\linewidth]{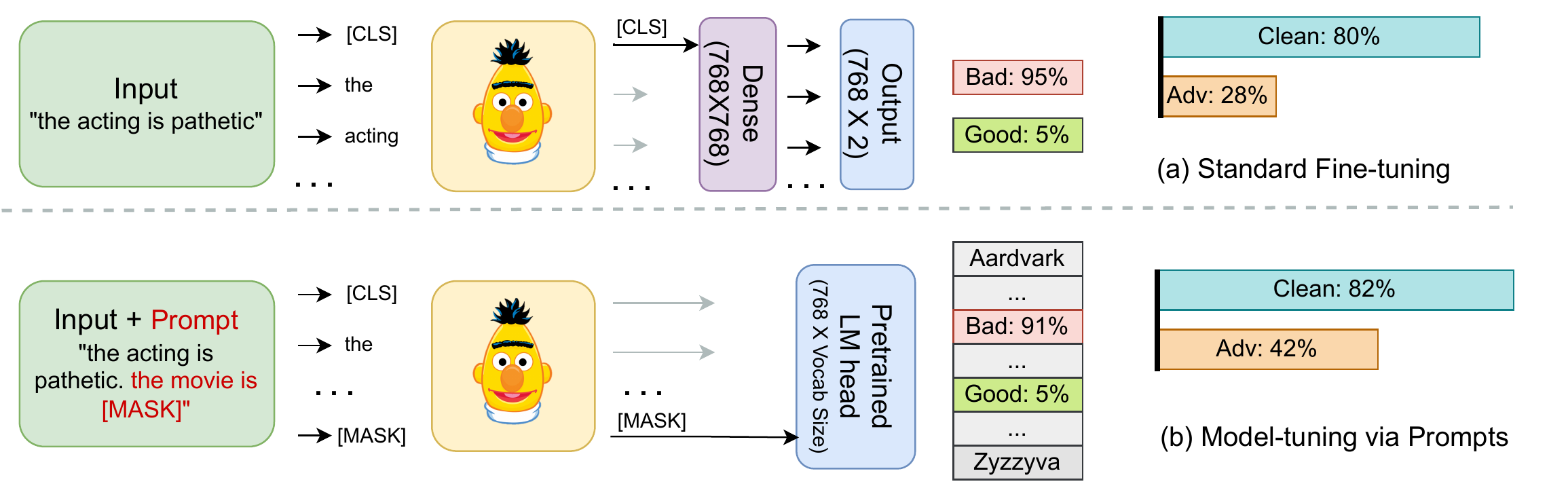}
  \caption{An illustration of (a) Standard Finetuning, and (b) Model-tuning via Prompts. The adjoining accuracy metrics correspond to a RoBERTa model trained on the BoolQ dataset.}
  \label{fig:mvp_illustration}
\end{figure*}

Pre-trained NLP models~\cite{devlin-etal-2019-bert,liu2019roberta} 
are typically adapted to downstream tasks 
by (i) appending a randomly initialized multi-layer perceptron 
to their topmost representation layer;
and then (ii) fine-tuning the resulting model
on downstream data (\linearFTns). 
More recently, work on large language models
has demonstrated comparable performance without fine-tuning, 
by just prompting the model with a prefix
containing several examples of inputs 
and corresponding target values~\citep{Brown2020Language}.
More broadly, prompting approaches 
recast classification problems 
as sequence completion (or mask infilling) tasks 
by embedding the example of interest into a prompt template.
The model's output is then mapped to a set of candidate answers to make the final prediction. 
Prompting has emerged as an effective strategy for
large-scale language models \citep{lester-etal-2021-power},
and its utility has also been demonstrated for masked language models
\citep{gao-etal-2021-making}.

While fine-tuned models perform well on in-distribution data, a growing body of work demonstrates that they remain brittle to adversarial perturbations~\citep{Jin_Jin_Zhou_Szolovits_2020,li-etal-2020-bert-attack,morris-etal-2020-textattack}.
Even small changes in the input text, 
such as replacement with synonyms~\citep{ebrahimi-etal-2018-hotflip},
and adversarial misspellings~\citep{ebrahimi-etal-2018-adversarial,pruthi-etal-2019-combating}
drastically degrade the accuracy of text classification models.
While prompting has become a popular approach 
for adapting pretrained models to downstream data,
little work has considered interactions 
between adaptation strategies and adversarial robustness.

In this work, \textbf{first}, we demonstrate surprising benefits 
of Model-tuning Via Prompts~(\ourPromptns) 
in terms of robustness to adversarial substitutions, as compared to the standard approach of
fine-tuning models with an MLP head (\linearFTns).
Notably, ~\ourPromptns, which does not utilize
any sort of adversarial training or prompt optimization/engineering
already yields higher adversarial robustness
compared to the state-of-the-art methods utilizing adversarial training 
by an average of $3.5\%$ across five datasets (classification, boolean question answering, and paraphrase detection), 3 models (BERT, RoBERTa, and GPT-2)
and four attacks (word and character-level substitutions)  (\S\ref{sec:all-results}). 
Moreover, we find that combining \ourPrompt with single-step adversarial training 
can further boost adversarial robustness,
resulting in combined robustness gains of more than $10\%$ 
over the baselines.
This happens without any loss in accuracy on unperturbed inputs, 
indicating how the objective of adversarial training couples well with \ourPromptns.

So far, prior works have not explored the idea of fine-tuning all the parameters of a model via prompts (we call this setup full-model full-data fine-tuning). We only see instances of (i) fine-tuning the full model via prompts in a few-shot setting \citep{gao-etal-2021-making}, or (ii) fine-tuning additional tunable parameters using prompts on top of a frozen model by utilizing the complete training set \citep{li-liang-2021-prefix}. We believe the idea of full-model full-data fine-tuning via prompts has not been used until now because clean accuracy improvements for \ourPrompt over \linearFT are negligible, and the robustness advantages of \ourPrompt were previously undiscovered. 

\textbf{Second}, 
we show that \ourPrompt as a method for classification 
is more (i) sample efficient, and (ii) has higher effective robustness 
than \linearFT (\S\ref{subsec:additional-advantages}).
That is,  \ourPrompt requires fewer training samples 
to achieve the same clean accuracy; 
and for any given clean accuracy, 
the robust accuracy of \ourPrompt is higher than \linearFTns.
Through ablation studies (\S\ref{subsec:ablation}),
we find that (i) adding multiple prompt templates
makes it harder to fool the model;
and (ii)  having multiple candidate answers has a small but positive impact
on the robustness.

\textbf{Third}, to explain our observations, we test a set of hypotheses
(\S\ref{sec:hypothesis}), including 
(i) \emph{random parameter vulnerability}---is 
adding a randomly initialized linear head 
the source of adversarial vulnerability for \linearFTns?;
(ii) \emph{pretraining task alignment}---can 
the gains in robustness be attributed to the alignment
between the fine-tuning and pretaining tasks in \ourPromptns?;
and (iii) \emph{semantically similar candidates}---are predictions
by \ourPrompt more robust because the candidate answer 
is semantically similar to the class label? 
Through experiments designed to test these hypotheses, 
we find that (i) in the absence of pretraining,
\ourPrompt and \linearFT have similar robustness performance,
supporting the hypothesis of pretraining task alignment; 
(ii) adding extra uninitialized parameters to \ourPrompt 
leads to a sharp drop in robustness,
whereas removing the dense $(768, 768)$ 
randomly initialized weight matrix from \linearFT
improves the robustness of the model significantly;
(iii) even random candidate answers such as `jack', 
and `jill' result in similar robustness gains, 
suggesting that when fine-tuning through prompts,
the choice of candidate answers is inconsequential
(in contrast, this choice is known to be crucial for few-shot classification).

\textbf{Fourth}, we perform a user study (\S\ref{sec:discussion_main}) to assess the validity of adversarial examples.
We find that human annotators were $23\%$ more likely to find adversarial examples to have been perturbed as opposed to clean examples. Moreover, humans achieved $11\%$ lower accuracy on adversarial examples as compared to clean examples with average confidence on the label of perturbed examples being $15\%$ lower.
This highlights that a large fraction of adversarial examples are already detected by humans, and often change the true label of the input, signifying that  \ourPrompt is more robust than crude statistics discussed in \S\ref{sec:all-results}. Future work will benefit from developing better evaluation strategies for the robustness of NLP models.

\textbf{Fifth}, going beyond adversarial robustness, we investigate the robustness gains of \ourPrompt over \linearFT on out-of-distribution (OOD) tasks. We find 
that \ourPrompt improves robustness by 2\%
across 5 different OOD sentiment analysis tasks (\S~\ref{app:ood_expts}).

In summary, we demonstrate that models tuned via prompts (\ourPromptns) are considerably more robust than the models adapted through the conventional approach of fine-tuning with an MLP head. Our findings suggest that practitioners adopt \ourPrompt as a means of fine-tuning, regardless of the training data size (few-shot or full data) and model capacity.

\section{Related Work}

\paragraph{Adversarial Attacks and Defenses}
Inspired by the brittleness of vision models to adversarial examples \citep{Szegedy2013IntriguingPO, Goodfellow2014ExplainingAH}, researchers have found similar vulnerabilities to  also exist in language models \citep{alzantot-etal-2018-generating,belinkov2018synthetic}. Unlike vision, the goal in NLP is to develop (i) semantically viable substitutions or deletions \citep{ebrahimi-etal-2018-hotflip}; (ii) character-level misspellings \citep{Zhang2015CharacterlevelCN,ebrahimi-etal-2018-adversarial,pruthi-etal-2019-combating}; or (iii) imperceptible homoglyphs \citep{boucher2022bad}.

The discovery of such adversarial examples span several tasks such as classification~\citep{Zhang2015CharacterlevelCN,alzantot-etal-2018-generating}, NMT~\citep{belinkov2018synthetic}, and question-answering~\citep{jia-liang-2017-adversarial}, but they are restricted to small models such as LSTMs and RNNs. Among others, \citet{Jin_Jin_Zhou_Szolovits_2020, li-etal-2020-bert-attack} show that despite producing massive gains on standard NLP benchmarks, BERT style pretrained models are susceptible to adversarial attacks when finetuned on downstream tasks. Subsequently, multiple works have attempted at developing fast and semantically meaningful attacks~\citep{li2018textbugger} and scalable defenses~\citep{wang-bansal-2018-robust,jia-etal-2019-certified,wang2021natural, si-etal-2021-better, Zhu2020FreeLB:} for masked language models. \citet{yang2022prompting} leverage prompts to generate adversarial examples that they train their model on using \linearFT. Despite these efforts, NLP models suffer a significant drop in robust accuracy, when compared to clean accuracy on the same task.

\paragraph{Prompting NLP Models} Prompting gained traction from GPT-3 \citep{Brown2020Language} where it was primarily used in the zero-shot and few-shot settings and required manual trials to increase performance. In the zero-shot setting, no labeled examples are provided to the model and the language model is kept frozen. The model needs to output its prediction using the prompt that is provided. Whereas, in the few-shot setting, a few task-specific labeled examples are also provided for the frozen model in addition to the prompt (also known as in-context learning)~\citep{rubin-etal-2022-learning, levine2022the}.
A lot of work has gone into improving the prompts that are used in the zero-shot and few-shot settings, including mining-based methods to automatically augment prompts \citep{jiang2020can}, gradient-based search \citep{shin-etal-2020-autoprompt}, using generative language models \citep{gao-etal-2021-making} and others \citep{hu-etal-2022-knowledgeable, schick-schutze-2021-just, schick-schutze-2021-exploiting}. In the full data setting, previous works have explored prompting via prompt tuning \citep{liu-etal-2022-p, li-liang-2021-prefix, qin-eisner-2021-learning, lester-etal-2021-power}~where the model is injected with additional tunable parameters. None of these works discuss the robustness advantages of prompting (especially in the adversarial context) when compared to standard fine-tuning approaches.

\paragraph{Robust Fine-tuning and Adaptation}
In the vision literature, prior works have also tried to use prompting to improve out-of-distribution robustness in the zero-shot and few-shot settings \citep{zhou2022conditional, zhou2022learning}.
\citet{kumar2022finetuning} observed that 
fine-tuning worsens the out-of-distribution (OOD) performance of models due to
the bias introduced via a randomly-initialized head on top of the CLIP model, and instead suggest a procedure (\LPFTns) that first fits the linear head and then finetunes the model.  Later works have shown that this ID/OOD performance trade-off could be mitigated by averaging model weights between the original zero-shot and fine-tuned model \cite{wortsman2022robust} and/or by finetuning using an objective similar to that used for pretraining \cite{goyal2022flyp}.  However, this work has been applied only to vision--language models, and secondly only deals with ``natural'' robustness evaluations
rather than the adversarial manipulations we consider here.

\section{Method}
\label{sec:method}


We consider the task of supervised text classification, where
we have a dataset $\mathcal{S} = \{x^{(i)}, y^{(i)}\}^n$, with $x^{(i)} \in \mathcal{X}$ and $y^{(i)} \in \{1,\dots,k\}$ for a $k$-class classification problem.  We train a classifier $f$ to predict $y$ based on input $x$. 
We follow the terminology by \citet{schick-schutze-2021-exploiting}.
The input $(x)$ can be decomposed as a sequence of words $\{x_1, x_2, \dots, x_l\}$, and the output $(y)$ is a positive integer, with each value corresponding to a particular class. The prompt template $(t)$ is the input string we append at the beginning or end of the input. For example, we may append the following template at the end of a movie review---"This movie is \mask". The candidate answers ($\mathcal{A}$) is a set of words corresponding to each class. For example, the positive sentiment class can have the following candidate answers---\{great, good, amazing\}.

\paragraph{Adversarial Attacks} We are concerned with perturbations to the input $x$
that change the model prediction.
In the case of adversarial attacks confined to synonym substitutions, we confine the model to searching for $\hat{x_i}$ in the synonym set of every word $x_i$ in the input. Whereas, in the case of character level substitution, we consider substitutions of characters that compose each $x_i$ in the input.

\subsection{Model-tuning Via Prompts (\ourPromptns)}
We present the overall pipeline of \ourPrompt in Figure~\ref{fig:mvp_illustration}(b), and describe individual components below.

\paragraph{Input Modification}
Consider a prompt template $t = {t_1, t_2, \dots \mask, \dots t_{m}}$. For any input $x$, the prompt input ($x_t$) can be constructed by appending the template at the beginning or end of the input. 
The final output is based on the most likely substitution
for the \mask token, as given by the language model. 
Typically, we use a set of prompt templates denoted by $\mathcal{T}$.

\paragraph{Inference}
For every class label, we have a set of candidate answers associated with it. During inference, we do the following: (i) for every class label, select the candidate corresponding to the largest logit value among its candidate set; (ii) take the mean of the logits corresponding to the selected candidates over all the templates to compute the final logit of the given class label; (iii) predict the class having the highest final logit.

\subsection{\ourPrompt + Single-step Adv}
\label{subsec:single-step-adv}
Based on the Fast Gradient Sign Method (FGSM) by \citet{Goodfellow2014ExplainingAH}, we perform single-step adversarial training. 
Note that the input tokens are discrete vectors, and hence it is not possible to perturb the inputs directly. Instead, we pass the inputs through the embedding layer of the model and then perform adversarial perturbations in the embedding space. 
We do not perturb the embeddings corresponding to the prompt tokens.
We find that performing single-step perturbations with the $\ell_2$ constraint leads to more stable training than in the $\ell_\infty$ norm ball, and use the same for all our experiments. Similar (but not equivalent) approaches have also been studied in literature~\cite{si-etal-2021-benchmarking}. 

\begin{table*}[t]
\centering
\scalebox{\tablescale}{
\begin{tabular}{@{}lccccccc@{}}
\toprule 
& \multicolumn{7}{c}{SST2}\\ \toprule
                    & \linearFT & \linearFT + Adv & Free LB++ & MADA & InfoBert & \ourPrompt & \ourPrompt + Adv \\\midrule
Clean Acc      & 93.6 \footnotesize $\pm$0.4 & 93.6 \footnotesize $\pm$0.6 & 94.0 \footnotesize $\pm$0.1 & 93.8 \footnotesize $\pm$0.4 & 94.0 \footnotesize $\pm$0.4 & 93.9 \footnotesize $\pm$0.7 & 93.8 \footnotesize $\pm$0.1 \\
TextFooler          & 40.2 \footnotesize $\pm$0.9 & 44.0 \footnotesize $\pm$1.2 & 43.4 \footnotesize $\pm$1.0 & 41.8 \footnotesize $\pm$0.5 & 43.6 \footnotesize $\pm$0.5 & \uline{46.9 \footnotesize {$\pm$0.5}} & \textbf{53.8 \footnotesize $\pm$0.7} \\
TextBugger          & 65.4 \footnotesize $\pm$0.3 & 68.5 \footnotesize $\pm$1.5 & 67.2 \footnotesize $\pm$0.6 & 66.1 \footnotesize $\pm$0.2 & 66.6 \footnotesize $\pm$1.8 & \uline{69.8 \footnotesize {$\pm$0.5}} & \textbf{71.7 \footnotesize $\pm$0.8} \\
BertAttack          & 70.3 \footnotesize $\pm$0.9 & 74.3 \footnotesize $\pm$0.8 & 76.2 \footnotesize $\pm$0.6 & 74.2 \footnotesize $\pm$0.2 & 76.1 \footnotesize $\pm$0.6 & \uline{78.1 \footnotesize {$\pm$0.9}} & \textbf{81.7 \footnotesize $\pm$0.7} \\
Misspellings        & 45.2 \footnotesize $\pm$1.1 & 49.3 \footnotesize $\pm$0.3 & 50.4 \footnotesize $\pm$1.1 & 45.4 \footnotesize $\pm$0.4 & 47.1 \footnotesize $\pm$0.4 & \uline{50.5 \footnotesize {$\pm$0.7}} & \textbf{54.9 \footnotesize $\pm$1.3} \\
\toprule
& \multicolumn{7}{c}{AG News}                                               \\\toprule
& \linearFT & \linearFT + Adv & Free LB++ & MADA & InfoBert & \ourPrompt & \ourPrompt + Adv \\ \midrule
Clean Acc   & 94.5 \footnotesize $\pm$0.4 & 94.4 \footnotesize $\pm$0.6 & 94.4 \footnotesize $\pm$0.7 & 94.1 \footnotesize $\pm$0.6 & 94.5 \footnotesize $\pm$0.9 & 94.3 \footnotesize $\pm$0.2 & 94.4 \footnotesize $\pm$0.8 \\
TextFooler          & 42.9 \footnotesize $\pm$0.7 & 47.7 \footnotesize $\pm$0.5 & 46.9 \footnotesize $\pm$1.6 & 44.3 \footnotesize $\pm$1.4 & 48.0 \footnotesize $\pm$2.2 & \uline{51.5 \footnotesize {$\pm$2.1}} & \textbf{62.7 \footnotesize $\pm$2.4} \\
TextBugger          & 61.8 \footnotesize $\pm$0.3 & 65.6 \footnotesize $\pm$0.8 & 65.5 \footnotesize $\pm$1.0 & 62.9 \footnotesize $\pm$0.5 & 65.6 \footnotesize $\pm$1.2 & \uline{68.7 \footnotesize {$\pm$0.7}} & \textbf{75.3 \footnotesize $\pm$1.6} \\
BertAttack          & 79.1 \footnotesize $\pm$1.3 & 81.1 \footnotesize $\pm$1.0 & 81.4 \footnotesize $\pm$0.9 & 80.4 \footnotesize $\pm$0.2 & 82.4 \footnotesize $\pm$1.2 & \uline{85.3 \footnotesize {$\pm$1.3}} & \textbf{88.2 \footnotesize $\pm$0.9} \\
Misspellings        & 76.8 \footnotesize $\pm$1.3 & 78.6 \footnotesize $\pm$0.8 & 80.1 \footnotesize $\pm$1.3 & 77.1 \footnotesize $\pm$0.4 & 80.4 \footnotesize $\pm$1.4 & \uline{82.7 \footnotesize {$\pm$0.7}} & \textbf{86.6 \footnotesize $\pm$0.6} \\
\toprule
& \multicolumn{7}{c}{BoolQ}                                               \\\toprule
& \linearFT & \linearFT + Adv & Free LB++ & MADA & InfoBert & \ourPrompt & \ourPrompt + Adv \\ \midrule
Clean Acc     & 80.6 \footnotesize $\pm$1.5 & 78.9 \footnotesize $\pm$1.2 & 80.6 \footnotesize $\pm$0.4 & 79.2 \footnotesize $\pm$0.9 & 81.5 \footnotesize $\pm$0.7 & 82.0 \footnotesize $\pm$0.6 & 81.1 \footnotesize $\pm$0.6 \\
TextFooler          & 28.2 \footnotesize $\pm$1.7 & 39.0 \footnotesize $\pm$0.7 & 37.2 \footnotesize $\pm$1.4 & 32.0 \footnotesize $\pm$0.3 & 38.0 \footnotesize $\pm$1.3 & \uline{42.9 \footnotesize {$\pm$0.5}} & \textbf{52.2 \footnotesize $\pm$1.6} \\
TextBugger          & 38.3 \footnotesize $\pm$1.0 & 44.4 \footnotesize $\pm$1.2 & 43.2 \footnotesize $\pm$1.0 & 41.1 \footnotesize $\pm$0.2 & 42.4 \footnotesize $\pm$1.5 & \uline{46.8 \footnotesize {$\pm$0.9}} & \textbf{56.7 \footnotesize $\pm$1.2} \\
BertAttack          & 48.1 \footnotesize $\pm$0.7 & 57.6 \footnotesize $\pm$1.3 & 57.3 \footnotesize $\pm$1.5 & 55.2 \footnotesize $\pm$0.4 & 57.4 \footnotesize $\pm$1.0 & \uline{61.5 \footnotesize {$\pm$1.2}} & \textbf{69.4 \footnotesize $\pm$1.5} \\
Misspellings        & 42.9 \footnotesize $\pm$1.0 & 47.4 \footnotesize $\pm$1.1 & 46.6 \footnotesize $\pm$1.1 & 45.2 \footnotesize $\pm$0.3 & 47.3 \footnotesize $\pm$1.2 & \uline{51.6 \footnotesize {$\pm$0.8}} & \textbf{59.7 \footnotesize $\pm$1.0} \\
\bottomrule
\end{tabular}
}
\caption{\textbf{Adversarial Robustness: } Performance of RoBERTa-base model on $3$ different datasets averaged over $3$ different seeds on a fixed test set of size 1000. The highest accuracies are bolded, and the second-best is underlined. 
We observe that models tuned via prompts (MVP) are the most robust while preserving (or improving) the clean accuracy.
}
\label{tab:acc-main}
\end{table*}
\section{Experimental Setup}
\label{sec:experiments}


\paragraph{Datasets and Models}
We perform our experiments on five different datasets---AG News~\citep{Zhang2015CharacterlevelCN} (4-class topic classification), SST-2~\cite{socher-etal-2013-recursive} (binary sentiment classification), BoolQ~\citep{clark-etal-2019-boolq} (boolean question answering), DBPedia14~\cite{dbpedia} (14-class topic classification), and MRPC~\cite{dolan-brockett-2005-automatically} (paraphrase detection). Results on DBPedia14 and MRPC are presented in Appendix~\ref{app:sec:full-results}.
All models are trained with the RoBERTa-Base~\cite{zhuang-etal-2021-robustly} backbone.
Experiments on GPT-2 and BERT-Base~\citep{devlin-etal-2019-bert} are included in 
Appendix~\ref{app:ext_expts}. Detailed information about training and attack hyperparameters is provided in Appendix~\ref{appendix:hyperparameters}.


\paragraph{Attack Strategies}
We perturb the inputs using the TextAttack library~\citep{morris2020textattack}. 
In particular, we use $1$ character-level attack and $3$ word-level attacks. Word-level attacks include the TextFooler~\citep{Jin_Jin_Zhou_Szolovits_2020}, TextBugger~\citep{li2018textbugger} that replace words with neighboring words based on counterfitted GloVe embeddings and BertAttack~\cite{li-etal-2020-bert-attack} that uses BERT to replace keywords with synonyms.\footnote{In line with previous benchmark~\cite{li-etal-2021-searching} we only use the word-substitution transformation in TextBugger.} 
For character-level attack, we use adversarial misspellings~\cite{pruthi-etal-2019-combating}. 
More details are in Appendix~\ref{app:sec:details}.

\paragraph{Baseline Methods}
We now describe the terminologies used to denote training schemes corresponding 
to various fine-tuning strategies.
\linearFT is the ``base'' model for classification via standard non-adversarial training, and is utilized by all the baselines. Given a pre-trained model, we perform downstream fine-tuning by adding an MLP layer to the output corresponding to \texttt{[CLS]} token as illustrated in Figure~\ref{fig:mvp_illustration}(a). This hidden representation is of size $768\times 1$. In the case of the BERT model, there is a single dense layer of dimension $768\times2$, whereas in the case of RoBERTa model, we have a two-layer MLP that is used to make the final prediction. 
\linearFT + Adv is 
is identical to the method used for adversarial training in Section~\ref{subsec:single-step-adv}, wherein
we perform adversarial perturbations in the embedding space of the \linearFT model, rather than \ourPromptns. 
To compare with state-of-art adversarial training-based defenses we consider
FreeLB++~\citep{li-etal-2021-searching} (free large batch adversarial training using projected gradient descent), InfoBERT~\citep{wang2021infobert} (information bottleneck regularizer to suppress noisy information), and AMDA~\citep{si-etal-2021-better}
(adversarial and
mixup data augmentation for creating new training examples via interpolation). We provide complete details pertaining to each baseline method in Appendix~\ref{app:baselines}.

\section{Results}
\label{sec:all-results}
\begin{table*}[t]
\centering
\scalebox{\tablescale}{
\begin{tabular}{lccllllll}
\toprule
 &  &  & \multicolumn{3}{c}{BoolQ} & \multicolumn{3}{c}{AGNews} \\
 \midrule
\textbf{Experiment} & \# Templates & Candidate & Clean & TFooler & TBugger & Clean & TFooler & TBugger \\
\midrule 
\linearFT & N/A & N/A & 80.6 \footnotesize $\pm$~1.5 & 28.2 \footnotesize $\pm$~1.7 & 38.3 \footnotesize $\pm$~1.0 & 94.5 \footnotesize $\pm$~0.4 & 42.9 \footnotesize $\pm$~0.7 & 61.8 \footnotesize $\pm$~0.3\\
\midrule
\multicolumn{1}{c}{} & {1} & Class Label & 81.9 \footnotesize $\pm$~0.8 & 35.9 \footnotesize $\pm$~0.2 & 44.6 \footnotesize $\pm$~0.5 & 94.6 \footnotesize $\pm$~0.4 & 48.6 \footnotesize $\pm$~1.1 & 67.3 \footnotesize $\pm$~1.1 \\
\multicolumn{1}{c}{} &{2} & Class Label & 82.3 \footnotesize $\pm$~0.2 & 37.4 \footnotesize $\pm$~0.3 & 46.4 \footnotesize $\pm$~0.5 & 94.5 \footnotesize $\pm$~0.6 & 50.8 \footnotesize $\pm$~1.6 & 67.8 \footnotesize $\pm$~0.5 \\
\multicolumn{1}{c}{} & {3} & Class Label & 82.1 \footnotesize $\pm$~0.3 & 40.8 \footnotesize $\pm$~1.5 & 49.5 \footnotesize $\pm$~1.1 & 94.2 \footnotesize $\pm$~0.2 & 48.4 \footnotesize $\pm$~3.4 & 66.2 \footnotesize $\pm$~1.1 \\

\multicolumn{1}{l}{\multirow{-4}{*}{\begin{tabular}[c]{@{}l@{}}Template \\ Expansion\end{tabular}}}
& {4} & Class Label & 82.0 \footnotesize $\pm$~0.6 & 42.9 \footnotesize $\pm$~0.5 & 49.8 \footnotesize $\pm$~1.6 & 94.3 \footnotesize $\pm$~0.2 & 51.4 \footnotesize $\pm$~2.0 & 68.7 \footnotesize $\pm$~0.7 \\\midrule
Candidate Exp. & {4} & {Multiple} & 81.6 \footnotesize $\pm$~1.2 & 46.1 \footnotesize $\pm$~1.6 & 53.0 \footnotesize $\pm$~0.7 & 93.6 \footnotesize $\pm$~0.4 & 54.0 \footnotesize $\pm$~0.7 & 69.8 \footnotesize $\pm$~0.3 \\
 \bottomrule
\end{tabular}
}
\caption{\textbf{Ablation Studies: }We study the impact of the number of candidate answers and prompt templates on adversarial performance of \ourPrompt(see~\S\ref{subsec:ablation}). `TFooler' and `TBugger' represent model robustness under
TextFooler and TextBugger attacks respectively. `Clean' represents model accuracy on original test data.  Additionally, we also assess the effect of including semantically similar answer candidates (see~\S\ref{sec:hypothesis}). All values are averaged over $3$ seeds.  
}
\label{tab:ablation-1}
\end{table*}

We first evaluate the impact of using \ourPrompt on the adversarial robustness of NLP models. 
For the task of Boolean question answering (BoolQ), 
we find that 
fine-tuning a RoBERTa model with an MLP head (\linearFTns) achieves an accuracy of $28.2\%$ 
on adversarial examples obtained through the TextFooler attack strategy (Table \ref{tab:acc-main}). 
Whereas, the corresponding accuracy for tuning the model via prompts (\ourPromptns) is $42.9\%$ which is a considerable improvement over \linearFTns. Additionally, \ourPrompt leads to more robust models compared to adversarial training baselines like \linearFT + Adv and InfoBERT that attain accuracies of $39.0\%$ and $38.1\%$ respectively. Further, \ourPrompt can be combined with adversarial training (\ourPrompt + adv), and doing so leads to an accuracy of  $52.2\%$ which is about a $10\%$ improvement over \ourPromptns, without any loss in clean performance. 

Similar to boolean question answering, the robustness advantages of \ourPrompt hold across the three tasks we examine. 
The individual performance statistics are detailed in~Table~\ref{tab:acc-main}. 
Overall, 
across four attack strategies, and three datasets, we 
report that \ourPrompt improves over 
\linearFT by $8\%$.
Remarkably, even in the absence of any adversarial training \ourPrompt 
achieves the state-of-the-art adversarial performance 
improving baseline adversarial training methods by $3.5\%$. Moreover, 
it can be coupled with single-step adversarial training, resulting 
in an overall $7\%$ improvement over state-of-art methods.
Lastly, 
the robustness benefits come only at a $2$x computation cost of 
standard training, as opposed to past works which need $5$--$10$x computation cost of standard training due to additional adversarial training. Results on BERT-Base are in Table~\ref{table:app:bert-full}.

\subsection{Sample Efficiency \& Effective Robustness}
\label{subsec:additional-advantages}

\begin{figure}
\centering
\begin{subfigure}[t]{0.23\textwidth}
\includegraphics[width=\linewidth]{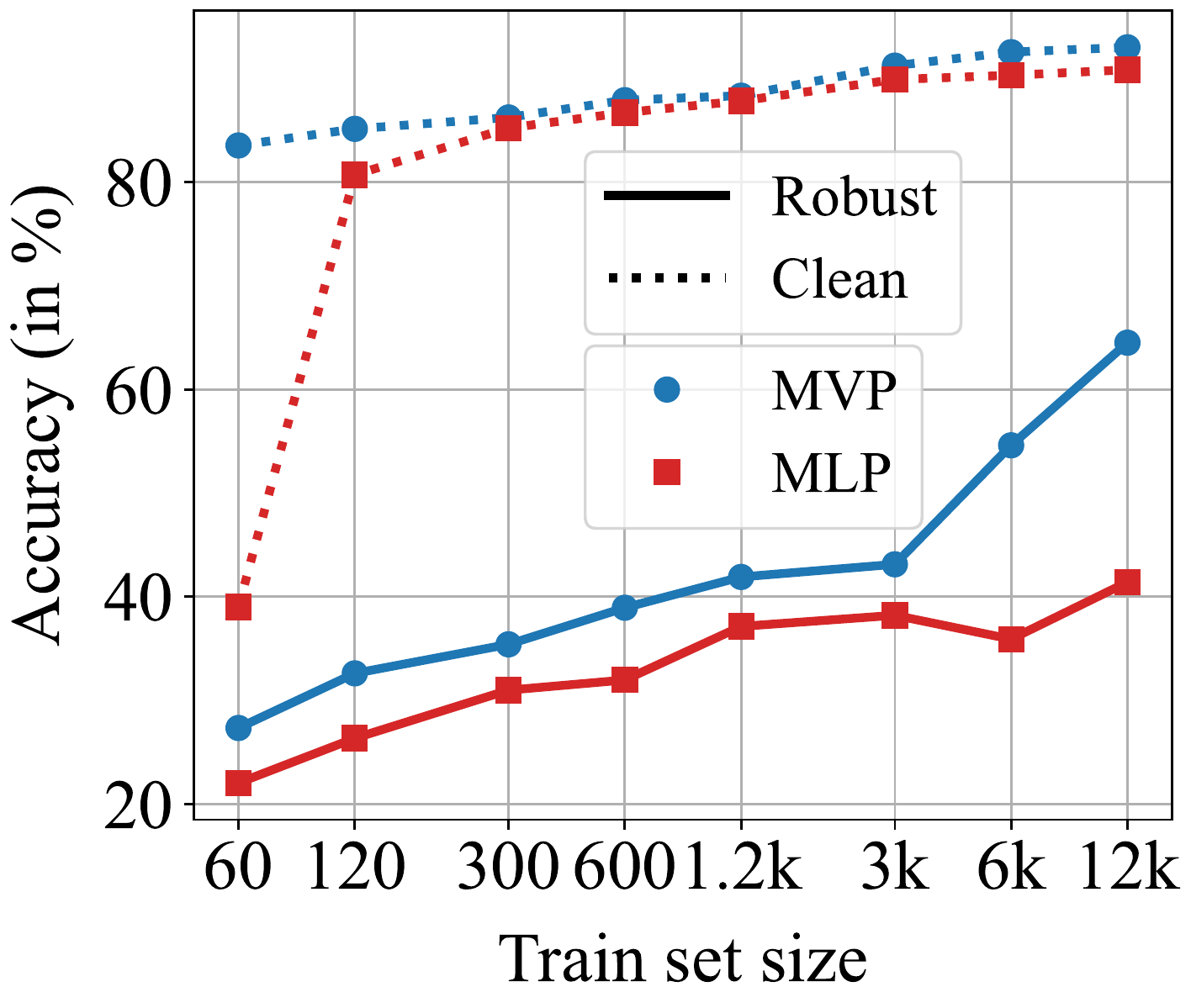}
  \label{fig:eff}
\end{subfigure}
\begin{subfigure}[t]{0.23\textwidth}
\includegraphics[width=\linewidth]{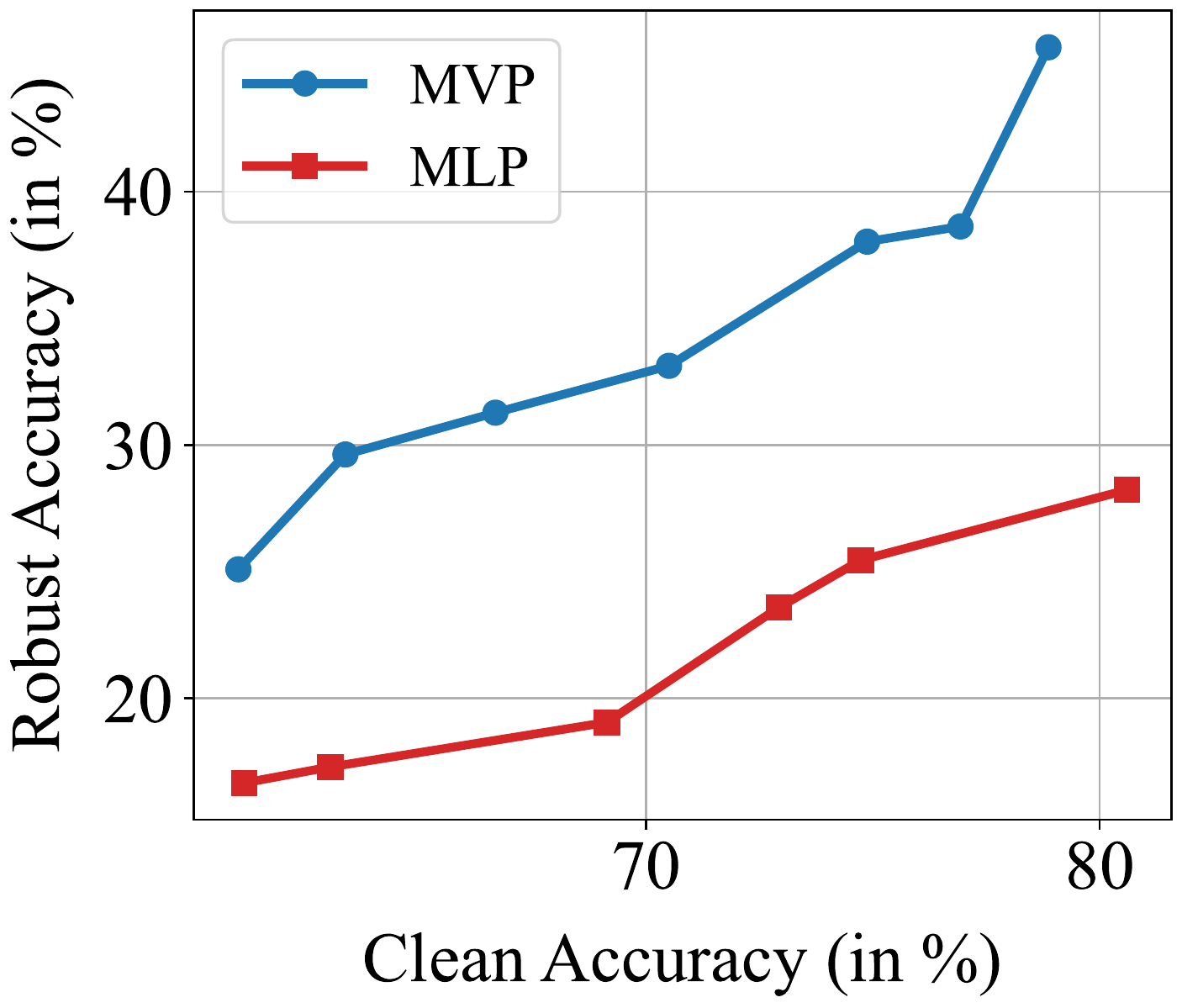}
\label{fig:scaling-curves}
\end{subfigure}
\caption{(a) \textbf{Sample Efficiency}: Clean and Robust Accuracy of RoBERTa-base model when trained using different data sizes of the AG News dataset. (b) \textbf{Effective Robustness}: Robust vs Clean Accuracy of RoBERTa-base model on the BoolQ dataset. 
We find that (a) \ourPrompt is more sample efficient as compared to \linearFT, and (b) \ourPrompt yields more robustness compared to \linearFT for the same clean accuracy
(see \S\ref{subsec:additional-advantages} for details).
}
\label{fig:scaling-curves}
\end{figure}

We investigate the sample efficiency and effective robustness
of \ourPrompt 
through experiments on the BoolQ and AG-News datasets using the RoBERTa-base model.
We train models on randomly sampled fractions of the dataset, ranging from $5\cdot10^{-4}$ to $0.1$.  

\paragraph{Sample Efficiency}
We compare the performance of \ourPrompt and \linearFT in low-data regimes. 
We find that \ourPrompt results in models are consistently more robust compared to models trained through \linearFT in the low data setups (see Figure~\ref{fig:scaling-curves}a). 
In fact, we 
observe that in extremely low resource case (only $60$ examples), 
it is hard to learn using \linearFT, 
but model trained through \ourPrompt performs exceedingly well. 
We note 
that the 
relative benefit of \ourPrompt over \linearFT peaks around $5$--$10\%$ of the data. Interestingly, the model trained through \ourPrompt requires only $5\%$ of samples to achieve similar robustness levels 
as models trained with \linearFT on the full dataset. 
In addition to robustness benefits,  
we find that \ourPrompt 
achieves considerably higher clean accuracy in low-data regimes (i.e., with $<200$ examples). 
Results on BoolQ are in \ref{app:subsec:addn_eff}.

\paragraph{Effective Robustness}
Effective robustness~\citep{taori2021measuring} measures the
robust accuracy 
of models that have the same clean accuracy. This can help determine which training time design decisions will be valuable when scaled up. We measure the effective robustness for models trained through
\ourPrompt and \linearFT
by training them on different data sizes.
We find that even when both \linearFT and \ourPrompt achieve the same clean accuracy, models trained through \ourPrompt are more robust (Figure~\ref{fig:scaling-curves}b). 
Results on AG News are presented in \ref{app:subsec:addn_eff}.

\subsection{Out of Distribution Robustness}
\label{app:ood_expts}
Going beyond adversarial 
robustness, we now perform 
experiments to assess the out-of-distribution robustness of \ourPromptns, \linearFTns, and \LPFTns. We use 5 sentiment classification datasets, namely SST2, Amazon Polarity~\cite{zhang2016characterlevel}, 
IMDb~\cite{maas-etal-2011-learning}, 
Movie Rationales~\cite{zaidan-eisner-piatko-2008:nips}, 
and Rotten Tomatoes~\cite{Pang+Lee:05a}. 
We fine-tune a Roberta model on 1000 examples of each of these datasets and evaluate all the datasets. 
Since all of these datasets are binary sentiment analysis datasets, we use the same template and candidate words across all the models (for both training and evaluation).
Based on our investigation, we see that across 5 different models (and 20 evaluations) the average accuracy for \ourPrompt (89.65\%) is 2\% more than \linearFT and 1.3\% more than that of LPFT.

\begin{table*}
\centering
\scalebox{0.65}{
\begin{tabular}{lcccccccccccccccc}
\toprule
Train v/s Eval & \multicolumn{3}{c}{SST2} & \multicolumn{3}{c}{Amazon Polarity} & \multicolumn{3}{c}{IMDb} & \multicolumn{3}{c}{Movie Rationales} & \multicolumn{3}{c}{Rotten Tomatoes} \\
\cmidrule(lr){2-4}
\cmidrule(lr){5-7}
\cmidrule(lr){8-10}
\cmidrule(lr){11-13}
\cmidrule(lr){14-16}
& MVP & MLP-FT & LPFT & MVP & MLP-FT & LPFT & MVP & MLP-FT & LPFT & MVP & MLP-FT & LPFT & MVP & MLP-FT & LPFT \\
\midrule
SST2 & 91.3 & 91.2 & 91.9 & 92.8 & 89.2 & 90.1 & 89.4 & 87.6 & 87.9 & 86.0 & 85.9 & 86.2 & 86.1 & 83.2 & 84.1 \\
Amazon Polarity & 90.9 & 88.5 & 89.0 & 92.9 & 92.9 & 93.4 & 92.0 & 91.2 & 91.1 & 85.9 & 83.9 & 84.2 & 86.1 & 83.3 & 84.5 \\
IMDb & 84.4 & 81.4 & 83.5 & 91.9 & 88.8 & 88.7 & 92.2 & 91.9 & 92.4 & 92.0 & 89.9 & 90.2 & 81.9 & 78.1 & 80.1 \\
Movie Rationales & 89.9 & 85.9 & 85.4 & 92.5 & 89.1 & 90.7 & 91.7 & 90.6 & 91.6 & 94.4 & 93.5 & 94.3 & 87.4 & 83.0 & 83.4 \\
Rotten Tomatoes & 92.4 & 92.1 & 92.9 & 92.6 & 89.5 & 90.4 & 90.9 & 88.6 & 90.2 & 86.4 & 83.9 & 84.7 & 87.2 & 87.1 & 87.2 \\
\midrule
Average & 89.8 & 87.8 & 88.5 & 92.5 & 89.9 & 90.7 & 91.3 & 90.0 & 90.6 & 89.0 & 87.4 & 87.9 & 85.7 & 83.0 & 83.9 \\
\bottomrule
\end{tabular}}
\caption{\textbf{OOD Robustness: }The results report the standard accuracy (in \%) of a model trained on the dataset in the left-most column, and evaluated on 5 different OOD datasets.
We see that across 5 different models (and 20 evaluations), the average accuracy for \ourPrompt (89.65\%) on OOD tasks is 2\% more than \linearFT and 1.3\% more than \LPFTns.}
\label{tab:ood-exp}
\end{table*}

These results 
in Table~\ref{tab:ood-exp}
show that \ourPrompt is 
superior to \linearFT 
and \LPFT for both 
adversarial and OOD robustness. 
In summary, \LPFT helps 
reduce the impact 
of random parameter vulnerability, but \ourPrompt additionally allows pre-training task alignment (the second hypothesis) hence resulting in superior performance and no fundamental trade-off be it OOD or adversarial robustness.

\subsection{Ablation Studies}
\label{subsec:ablation}
\paragraph{Number of Candidate Answers}
A larger candidate answer set
is shown to improve clean performance in the few-shot setting \citep{hu-etal-2022-knowledgeable}. 
Here, we investigate the impact of 
the size of the candidate answer set on the adversarial performance of models tuned via prompts.
The adversarial accuracy of the model with a single candidate answer is $42.9\%$, and it increases to $46.2\%$ upon using an answer set comprising $4$ candidates.\footnote{Details about candidates and templates are in Appendix \ref{app:sec:temp-candidates}}  
These results correspond to the RoBERTa-base model on BoolQ dataset against adversarial perturbations from the TextFooler attack. Overall, we observe an improvement of $1.0$--$3.5\%$ in adversarial accuracy when we use a larger candidate set across different settings (Table~\ref{tab:ablation-1}). A more detailed analysis of the same with a single prompt template is provided in Appendix~\ref{app:enemble-candidates}.


\paragraph{Number of Prompt Templates}
Another design choice 
that we consider 
is the number of prompt templates used for prediction. 
We conjecture that the 
adversary may find 
it difficult to flip the model prediction 
when we average logits across multiple templates. 
To evaluate this, we train \ourPrompt with different number of prompt templates (ranging from 1 to 4), and compare the adversarial robustness. 
We 
notice a steady improvement in the adversarial accuracy 
as we increase the number of templates 
which supports our initial conjecture (see Table \ref{tab:ablation-1}).
While increasing the number of templates improves the robustness of the downstream model, \ourPrompt achieves large robustness gains even with a single template (compared to \linearFTns). Hence, using multiple prompt templates is not the fundamental reason for the improved robustness of \ourPromptns. Further, in order to assess the impact of the `choice' of prompt templates used, we perform a more details analysis on the impact of prompt tuning for adversarial robustness of \ourPrompt in Appendix~\ref{app:prompt-tuning}. We find that even empty or random templates perform nearly similar to well-crafted prompts, and retain the robustness advantages of \ourPrompt over \linearFTns.



\begin{figure*}[t]
    \centering
    \vspace{-0.5in}
    \includegraphics[width=0.85\linewidth]{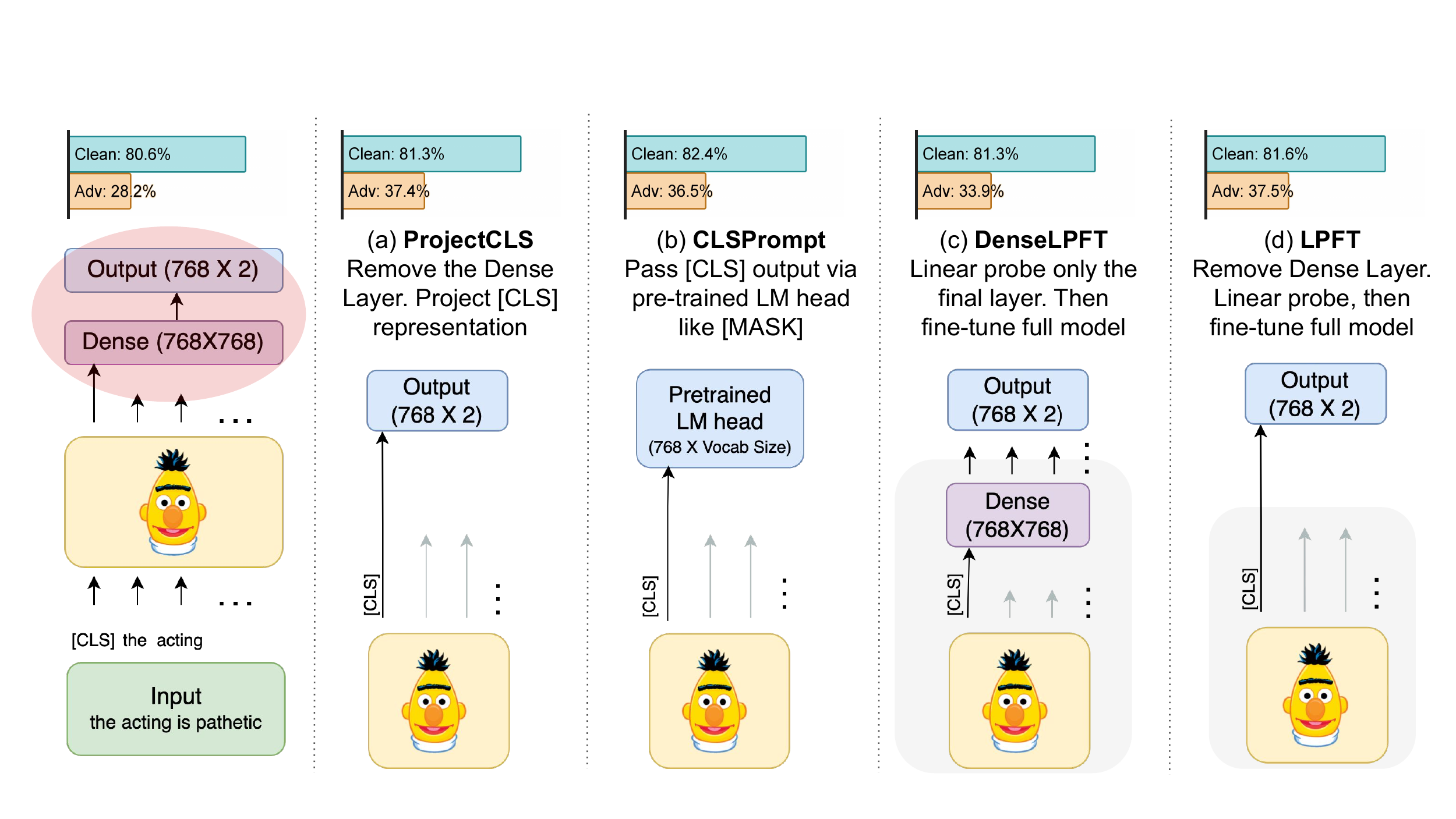}
    \caption{Various model tuning strategies for RoBERTa model trained on the BoolQ dataset. The corresponding clean and robust accuracies (under TextFooler attack) are also shown above each model paradigm. The left-most diagram shows the standard fine-tuning paradigm of \linearFT, and each subsequent column modifies the architecture, helping us confirm the hypothesis that randomly initialized parameters are a cause of vulnerability.}
    \label{fig:random_parameter}
\end{figure*}

\begin{table*}[t]
\centering
\scalebox{\tablescale}{
\begin{tabular}{llllllll}
\toprule
 &  & \multicolumn{3}{c}{\textbf{BoolQ}} & \multicolumn{3}{c}{\textbf{AGNews}}  \\
 \midrule
\textbf{Hypothesis} & Setting & Clean & TFooler & TBugger & Clean & TFooler & TBugger  \\
\midrule
& \linearFT & 80.6 \footnotesize $\pm$~1.5 & 28.2 \footnotesize $\pm$~1.6 & 38.3 \footnotesize $\pm$~1.0 & 94.5 \footnotesize $\pm$~0.4 & 42.8 \footnotesize $\pm$~0.7 & 61.8 \footnotesize $\pm$~0.3  \\
\midrule
 & \sparseCLS & 81.3 \footnotesize $\pm$~0.5 & 37.4 \footnotesize $\pm$~1.2 & 45.6 \footnotesize $\pm$~1.2 & 93.7 \footnotesize $\pm$~0.4 & 46.7 \footnotesize $\pm$~1.3 & 65.2 \footnotesize $\pm$~3.3  \\
 & \CLSPrompt & 82.4 \footnotesize $\pm$~0.3 & 36.5 \footnotesize $\pm$~0.4 & 46.0 \footnotesize $\pm$~1.2 & 94.7 \footnotesize $\pm$~0.2 & 47.2 \footnotesize $\pm$~1.9 & 66.7 \footnotesize $\pm$~2.0  \\
  & \denseLPFT & 81.3 \footnotesize $\pm$~0.4 & 33.9 \footnotesize $\pm$~1.4 & 42.6 \footnotesize $\pm$~1.2 & 94.5 \footnotesize $\pm$~0.6 & 44.2 \footnotesize $\pm$~0.8 & 64.5 \footnotesize $\pm$~1.1  \\
\multirow{-5}{*}{Random Parameter} & \sparseLPFT & 81.6 \footnotesize $\pm$~1.2 & 37.5 \footnotesize $\pm$~1.1 & 46.4 \footnotesize $\pm$~1.2 & 94.5 \footnotesize $\pm$~0.1 & 46.5 \footnotesize $\pm$~0.9 & 67.2 \footnotesize $\pm$~1.0  \\
\midrule
 & Untrained \ourPrompt & 67.5 \footnotesize $\pm$~0.9 & 11.7 \footnotesize $\pm$~2.7 & 14.9 \footnotesize $\pm$~2.7 & 90.1 \footnotesize $\pm$~0.8 & 12.2 \footnotesize $\pm$~2.9 & 20.6 \footnotesize $\pm$~2.2  \\
\multirow{-3}{*}{Task Alignment} & Untrained \linearFT & 67.0 \footnotesize $\pm$~0.6 & 14.8 \footnotesize $\pm$~4.3 & 17.5 \footnotesize $\pm$~1.1 & 89.5 \footnotesize $\pm$~0.4 & 13.4 \footnotesize $\pm$~1.2 & 19.4 \footnotesize $\pm$~0.8  \\
\midrule
Candidate Semantics & {Random (\ourPromptns)} & 80.9 \footnotesize $\pm$~0.3 & 42.1 \footnotesize $\pm$~0.4 & 48.1 \footnotesize $\pm$~2.2 & 93.4 \footnotesize $\pm$~0.3 & 50.3 \footnotesize $\pm$~1.2 & 68.3 \footnotesize $\pm$~0.3 \\
\bottomrule

\end{tabular}
}
\caption{Adversarial performance of RoBERTa for experiments corresponding to the random parameter vulnerability and task alignment hypotheses averaged over 3 seeds (\S\ref{sec:hypothesis}). 
`TFooler' and `TBugger' represent model robustness under TextFooler and TextBugger attacks respectively. `Clean' represents model accuracy on original test data.}
\label{tab:acc-untrained}
\end{table*}

\section{Why Does MVP Improve Robustness?}
\label{sec:hypothesis}
We test three hypotheses to explain the robustness gains achieved by \ourPrompt compared to \linearFT in the context of adversarial attacks.

\paragraph{Random Parameter Vulnerability}
One plausible explanation 
for the observed 
adversarial vulnerability of
\linearFT is the randomly-initialized linear head 
used for downstream classification.
The intuition behind this effect is that \emph{fine-tuning a set of randomly-initialized parameters may lead to feature distortion of the pretrained model} as is demonstrated in \citet{kumar2022finetuning}. This phenomenon has also been observed in CLIP models~\citep{Radford2021LearningTV}, 
where the authors found that fine-tuning the model using a randomly initialized linear prediction head reduces the out-of-distribution robustness of the model. The phenomenon is unexplored in the context of adversarial robustness. We study this effect through three experiments.



1. \underline{\sparseCLSns:}~~ First, we reduce the number of random parameters by removing the dense layer 
of weights ($768\times768$ parameters)
from the standard MLP architecture. We call this \sparseCLSns,
and only use 
a projection layer of dimensions $768\times C$ parameters, with
$C$ being the number of classes (see Figure~\ref{fig:random_parameter}(a)). 
We find that 
\sparseCLS is on average $\sim8\%$ more robust than \linearFT which suggests that reducing the number of randomly initialized parameters helps to increase model robustness (see Table~\ref{tab:acc-untrained}).

2. \underline{\CLSPromptns:}~~ Second, we train another model, \CLSPromptns, where instead of 
using the probabilities corresponding to the \texttt{[MASK]} token as in \ourPromptns, 
we use the probabilities of the candidate answers corresponding to the \texttt{[CLS]} token (see Figure~\ref{fig:random_parameter}(b)). 
The key difference between \CLSPrompt and \linearFT is that there are no randomly initialized MLP parameters in \CLSPromptns, and we 
use the probabilities corresponding to the candidate answers, instead of projecting the representations with new parameters. From Table \ref{tab:acc-untrained}, we observe that \CLSPrompt is once again on average $\sim8\%$  more robust than \linearFT  
which provides strong evidence in favor of our hypothesis 
of random parameter vulnerability.

3. \underline{\sparseLPFTns} (linear probe, then fine-tune): For our third experiment, we train two new models namely \sparseLPFT and \denseLPFT (see Figure~\ref{fig:random_parameter}(c,d)). 
In both these models, we do the following: (i) fit a logistic regression to the hidden states corresponding to the \texttt{[CLS]} token (linear probing);
(ii) initialize the final layer of the classification head with the learned $768\times C$ (where $C$ is the number of classes) matrix of the fitted logistic regression model; and (iii) fine-tune the whole model as in \linearFTns. 
The only difference between \sparseLPFT and \denseLPFT is that \denseLPFT has an additional randomly initialized dense layer of dimensions $768\times768$ unlike \sparseLPFTns.
In contrast to \citet{kumar2022finetuning}, we test \LPFT against adversarial manipulations. We note from Table \ref{tab:acc-untrained} that \denseLPFT is more robust than \linearFT (by over 10\%) but it demonstrates lower robustness as compared to \sparseLPFT (by over 2\%). This provides further evidence that randomly initialized parameters add to the vulnerability.

\paragraph{Pretraining Task Alignment}
The task of mask infilling aligns more naturally with the pretraining objective of the language model and we posit that finetuning via mask infilling as in \ourPrompt results in robustness gains.
To test this hypothesis, we use an untrained RoBERTa model, and measure the clean accuracy and robustness of \ourPrompt and \linearFT models. 
We observe that in the absence of pre-training, \ourPrompt trained with a single template does not achieve any additional robustness over the baseline, and in fact, \linearFT performs better than \ourPrompt (Table \ref{tab:acc-untrained}) whereas in the presence of pre-training, \ourPrompt outperforms \linearFT (Table \ref{tab:ablation-1}) in all the settings. Note that this does not contradict the hypothesis about vulnerability due to randomly-initialized parameters, as that hypothesis only applies for pretrained models. 
This suggests that the alignment 
of \ourPrompt with the pre-training task 
is crucial for adversarial robustness on downstream task.

\paragraph{Semantically Similar Candidates}
We question whether the improvement in robustness can also be attributed to the semantic relatedness
between candidate answers and the class labels. 
To answer this question, we change the candidate answers to random proper nouns (`jack', `john', `ann', `ruby') for the 4-class classification problem of AG-News and (`jack', `john') for the 2-class classification task of BoolQ. All of these words are unrelated to the class labels.
%
We find that irrespective of whether we use semantically related candidates or not, the robust accuracy of the model is within 1\% of each other, thereby implying that using semantically similar candidates is not a factor behind the robustness gains of \ourPrompt (Table~\ref{tab:acc-untrained}).
While the choice of candidate answers is crucial in the pre-train, prompt, and predict paradigm \citep{hu-etal-2022-knowledgeable}, it is irrelevant in the pre-train, prompt, and finetune paradigm. With sufficient fine-tuning over the downstream corpus, a model can learn to associate any candidate word with any class, irrespective of its semanticity.

However, one may wonder why using `random' candidate words doesn't hurt the model robustness, since this also leads to modifying a `parameter' in the model's embedding space, which was initially uncorrelated to the class label. We analyze this question in detail in Appendix~\ref{app:dummy-candidates} and conclude that the main reason for the preserved robustness is the `pre-training task hypothesis' and the fact that the modified word embeddings have a much smaller dimension of size 768 x C (where C is the number of candidate words), as opposed to modifying a dense layer.

\section{Human Study}
\label{sec:discussion_main}
We conduct a human study to assess the viability of the adversarial attacks. More specifically, we provide machine learning graduate students $250$ input examples and ask the following questions: (a) What is the perceived label of the sentence; (b) What is their confidence about this label; and (c) Was this sentence adversarially manipulated?
We use the BoolQ dataset and strictly instruct our annotators 
to not use any external knowledge but the context of the given passage only. We use samples that were successfully attacked by TextFooler for \ourPrompt + Adv model with a RoBERTa backbone. As a control for the study, we provide the original sentence rather than the adversarially perturbed one $33\%$ times. The underlying model achieves a clean accuracy of $81.7\%$ and a robust accuracy of $54.0\%$. 

We find that human annotators identify $29\%$ of adversarial examples to be perturbed as opposed to only $6\%$ of clean examples. Moreover, we also discover that humans achieved $11\%$ lower accuracy
on adversarial examples as compared to clean examples
($85\% \to 74\%$) with average confidence on the label
of perturbed examples being $15\%$ lower ($90\% \to 75\%$). This study highlights that a fraction of adversarial attacks either manipulate the input so significantly that it is easily detectable, or change the label, signifying that \ourPrompt is more robust than what crude statistics suggest in \S\ref{sec:all-results}. Details related to the human study are available in Appendix \ref{app:sec:human-study}.

\section{Conclusion}
In this work, we benchmark the robustness of language models when adapted to downstream classification tasks through prompting. Remarkably, model tuning via prompts---which does not utilize any sort of adversarial training or prompt engineering---already outperforms the state-of-the-art methods in adversarially robust text classification by over $3.5\%$ on average. 
Moreover, we find that \ourPrompt is sample efficient and also exhibits high \emph{effective} robustness as compared to the conventional approach of fine-tuning with an MLP head (\linearFTns). 
We find that the lack of robustness in baseline methods can largely be attributed to the lack of alignment between pre-training and finetuning task, and the introduction of new randomly-initialized parameters.

\section{Limitations}
This work considers models that are under 1B parameters in size. While larger models are becoming popular in the NLP community, developing practical attacks that scale to such large models is an extremely challenging task. For instance, for the evaluation considered in this paper, each attack takes approximately a day on a single A6000 GPU to run (across multiple seeds of the model). Furthermore, the scope of our work is limited to tasks where fine-tuning with an MLP head is commonplace. This includes boolean question answering, sentence classification, and paraphrase detection tasks. Finally, using multiple templates for \ourPrompt comes with a trade-off with latency which is discussed in Appendix~\ref{app:latency}.

\paragraph{Broader Impact} Our work does not pose any immediate negative impacts to society, except for the carbon emissions owing to the training and evaluation of big models. We emphasize that the  adversarial robustness conferred via \ourPrompt is a desirable property for deployed systems, and our work contributes towards making NLP models more reliable and safe when deployed in real-world settings.

\section*{Acknowledgements}
We thank the Pittsburgh weather that kept us from doing anything but work on this draft.
PM is supported by funding from the DARPA GARD
program.
ZL acknowledges the NSF (FAI 2040929 and IIS2211955) Amazon AI, UPMC, Highmark Health, Abridge, Ford, Mozilla, the PwC Center, the Block Center, the Center for Machine Learning and Health, and the CMU Software Engineering Institute (SEI) via Department of Defense contract FA8702-15-D-0002, for their generous support of ACMI Lab’s research.
DP is grateful for the support of Adobe Inc., Google and 
Kotak IISc AI-ML Centre (KIAC). 

\bibliography{anthology,paper}
\bibliographystyle{acl_natbib}

\clearpage
\appendix

\section*{\centering Supplementary Material\\ Model-tuning Via Prompts Makes NLP Models More Robust}

\section{Candidate Answers \& Prompt Templates}
\label{app:sec:temp-candidates}
We enumerate all the prompt templates and candidate answers used for our experiments on \ourPromptns. Templates beginning with the \texttt{[SEP]} token are appended at the end of the input otherwise they precede the input. Note that we remove the \texttt{[SEP]} token and then append the template to the input. The \texttt{[SEP]} token is just used as an indicator for appending the template to the input.
Note that since Causal Language models are not bidirectional, for GPT-2 experiments, all the prompt templates will be appended at the end of the input.
\paragraph{AG News}
The prompt templates used for MLMs:
\begin{enumerate}
    \item A \texttt{[MASK]} news  
    \item \texttt{[SEP]} This topic is about \texttt{[MASK]}
    \item Category : \texttt{[MASK]}  
    \item \texttt{[SEP]} The category of this news is \texttt{[MASK]} 
\end{enumerate}
The prompt templates used for GPT-2 are:
\begin{enumerate}
    \item \texttt{[SEP]} This topic is about \texttt{[MASK]}
    \item \texttt{[SEP]} The category of this text is \texttt{[MASK]}
    \item \texttt{[SEP]} Category : \texttt{[MASK]}
    \item \texttt{[SEP]} This is a news from \texttt{[MASK]} 
\end{enumerate}
The candidate answers used are the same as the class labels, namely---politics, business, sports, and technology---for all the experiments except the larger candidate set ablation study. For that ablation, we use the following candidate answer set:
\begin{enumerate}
    \item \{politics, world, government, governance\}
    \item \{sports, competition, games, tournament\}
    \item \{business, corporation, enterprise, commerce\}
    \item \{technology, science, electronics, computer\}
\end{enumerate}

\paragraph{BoolQ}
The prompt templates used for MLMs are:
\begin{enumerate}
    \item Answer to the question is \texttt{[MASK]}
    \item \texttt{[SEP]} \texttt{[MASK]}
    \item I think \texttt{[MASK]}
    \item \texttt{[SEP]} The answer is \texttt{[MASK]}
\end{enumerate}
The prompt templates used for GPT-2 are the same as above except every template is appended to the end of the input.
As in AG News, the candidate answers used are the same as the class labels, namely false and true, except when performing the larger candidate set experiment, in which case we use the following candidate answer set:
\begin{enumerate}
    \item \{false, wrong, incorrect, invalid\}
    \item \{true, correct, valid, accurate\}
\end{enumerate}

\paragraph{SST-2}
The prompt templates used for MLMs are:
\begin{enumerate}
    \item Sentiment of the statement is \texttt{[MASK]} . 
    \item \texttt{[SEP]} \texttt{[MASK]} 
    \item This is a \texttt{[MASK]} statement 
    \item \texttt{[SEP]} The statement is \texttt{[MASK]} 
\end{enumerate}

Similar to AG News and BoolQ, we use the class labels (i.e., negative and positive) as the candidate answers.

\paragraph{DBPedia14}
The prompt templates used for MLMs are:
\begin{enumerate}
    \item Content on \texttt{[MASK]}
    \item \texttt{[SEP]} This topic is about \texttt{[MASK]}
    \item Category : \texttt{[MASK]}
    \item \texttt{[SEP]} The content is about \texttt{[MASK]}
\end{enumerate}
The candidate answers used are:\\
\{0: `company', 1: `education', 2: `artist', 3: `athlete', 4: `office', 5: `transportation', 6: `building', 7: `nature', 8: `village', 9: `animal', 10: `plant', 11: `album', 12: `film', 13: `writing'\}
\paragraph{MRPC}
The prompt templates used for MLMs:
\begin{enumerate}
    \item The two sentences are \texttt{[MASK]}
    \item \texttt{[SEP]} First sentence is \texttt{[MASK]} to second sentence
    \item Two \texttt{[MASK]} sentences
    \item \texttt{[SEP]} The two sentences have \texttt{[MASK]} meanings
\end{enumerate}
The candidate answers used are:\\
\{0: `different', 1: `equivalent'\}

\begin{table*}[ht]
\centering
\scalebox{\tablescale}{
\begin{tabular}{@{}lllllll@{}}
\toprule
 & \multicolumn{6}{c}{GPT2} \\ \midrule
 & \multicolumn{3}{c}{BoolQ} & \multicolumn{3}{c}{AG News} \\ \midrule
 & \multicolumn{1}{c}{Clean Acc} & \multicolumn{1}{c}{TextFooler} & \multicolumn{1}{c}{TextBugger} & \multicolumn{1}{c}{Clean Acc} & \multicolumn{1}{c}{TextFooler} & \multicolumn{1}{c}{TextBugger} \\
 \midrule
\linearFT & 61.0 \footnotesize $\pm$ 2.1 & 20.2 \footnotesize $\pm$ 0.6 & 24.9 \footnotesize $\pm$ 1.4 & 93.7 \footnotesize $\pm$ 0.2 & 27.6 \footnotesize $\pm$ 1.2 & 58.2 \footnotesize $\pm$ 0.9 \\
\linearFT+Adv & 60.5 \footnotesize $\pm$ 0.4 & 22.0 \footnotesize $\pm$ 1.1 & 31.8 \footnotesize $\pm$ 1.8 & 92.4 \footnotesize $\pm$ 0.3 & \uline{39.6 \footnotesize {$\pm$ 0.5}} & 61.3 \footnotesize $\pm$ 0.7 \\
\ourPrompt & 72.5 \footnotesize $\pm$ 1.0 & \uline{28.7 \footnotesize {$\pm$ 1.6}} & \uline{38.3 \footnotesize {$\pm$ 1.6}} & 93.8 \footnotesize $\pm$ 0.3 & 31.4 \footnotesize $\pm$ 0.5 & \uline{61.0 \footnotesize {$\pm$ 0.8}} \\
\ourPrompt+Adv & 71.8 \footnotesize $\pm$ 0.8 & \textbf{30.1 \footnotesize $\pm$ 0.6} & \textbf{41.2 \footnotesize $\pm$ 0.8} & 93.7 \footnotesize $\pm$ 0.3 & \textbf{44.0 \footnotesize $\pm$ 0.2} & \textbf{64.4 \footnotesize $\pm$ 1.2} \\ \bottomrule
\end{tabular}}
\caption{Adversarial Robustness results on BoolQ and AG News dataset using GPT-2 model. All experiments are run on 3 different seeds and the performance is reported over a fixed test set of size 1000. The best-performing robust accuracies are bolded and the second best robust accuracies are underlined.}
\label{tab:gpt-acc}
\end{table*}
\section{Baseline Methods and Attacks}

\subsection{Baselines}
\label{app:baselines}
We describe training schemes corresponding 
to various fine-tuning strategies
below.

\textbf{\linearFT}: This is the ``base'' model for classification via standard non-adversarial training and is utilized by all the baselines discussed in this section. Given a pre-trained model, we perform downstream fine-tuning by adding an MLP layer to the output corresponding to \texttt{[CLS]} token as illustrated in Figure~\ref{fig:mvp_illustration}(a). This hidden representation is of size $768\times 1$. In the case of the BERT model, there is a single dense layer of dimension $768\times2$, whereas in the case of RoBERTa model, we have a two-layer MLP that is used to make the final prediction. 

\textbf{\linearFT + Adv}:
~This is identical to the method used for adversarial training in Section~\ref{subsec:single-step-adv}, wherein
we perform adversarial perturbations in the embedding space of the \linearFT model, rather than \ourPromptns. 

\textbf{FreeLB++}~\citep{li-etal-2021-searching}: 
Free Large-Batch (FreeLB) adversarial training~\citep{Zhu2020FreeLB:} performs multiple Projected Gradient Descent (PGD) steps to create adversarial examples, 
and simultaneously accumulates parameter gradients which are then used to update the model parameters (all at once).
FreeLB++ improves upon FreeLB by increasing the number of adversarial training steps to 10 and the max adversarial norm to 1. 

\textbf{InfoBERT}~\citep{wang2021infobert}: InfoBERT uses an Information Bottleneck regularizer to suppress noisy information that may occur in adversarial attacks. Alongside, an `anchored feature regularizer' tries to align local stable features to the global sentence vector. Together, this leads to improved generalization and robustness. InfoBERT can additionally be combined with adversarial training (we use Free LB++ for this purpose).
    
\textbf{AMDA}~\citep{si-etal-2021-better}: Adversarial and
Mixup Data Augmentation (AMDA) improves robustness to adversarial attacks by increasing the number of adversarial samples seen during training. This method interpolates training examples in their embedding space to create new training examples. The label assigned to the new example is the linear interpolation of the one hot encodings of the original labels.

\subsection{Attack Details}
\label{app:sec:details}
In the main paper, we evaluated our method on three popular word substitution attacks and one character-level attack. 
These included the TextFooler, TextBugger and BertAttack attack strategies. TextFooler and TextBugger are word substitution attacks that replace words with “similar” neighboring words (where similarity is based on counterfitted GloVe embeddings). TextFooler greedily searches in a large set of neighbors (in the embedding space) for each word, so long as they satisfy some constraints on embedding similarity and sentence quality. An additional constraint requires the substituted word to match the POS of the original word. TextBugger, on the other hand, restricts the search space to a small subset of neighboring words and only uses sentence quality as a constraint. To control the amount of change made by an attack, we limit the adversary to perturbing a maximum of 30\% words in the AG News dataset and 10\% in all other datasets. We do not modify any other constraints (such as the query budget) and run the attacks on 1000 examples from the test set.
We also evaluate on one character-level, and another word substitution attack. For character-level attack, we use the adversarial misspellings attack introduced by \citet{pruthi-etal-2019-combating}, and we additionally evaluate the popular BertAttack~\cite{li-etal-2020-bert-attack}.

\section{Extended Experiments on Adversarial Robustness}
\label{app:ext_expts}

\subsection{Results on Additional Datasets and Models}
\label{app:sec:full-results}
\begin{table*}[t]
\centering
\scalebox{\tablescale}{
\begin{tabular}{@{}lccccc@{}}
\toprule
             & \multicolumn{5}{c}{DBPedia}                                               \\\toprule
             & Clean Acc & TextFooler   & TextBugger   & BertAttack  & Misspellings       \\\midrule
\linearFT       & 97.3\footnotesize $\pm$0.7     & 43.8\footnotesize $\pm$1.5   & 68.7\footnotesize $\pm$0.9   & 72.4\footnotesize $\pm$1.2  & 65.7\footnotesize $\pm$1.3   \\
\linearFT + Adv & 97.2\footnotesize $\pm$0.4     & 56.1\footnotesize $\pm$0.2   & 76.4\footnotesize $\pm$0.3   & 78.3\footnotesize $\pm$0.6  & 72.2\footnotesize $\pm$0.7   \\
\ourPrompt       & 97.0\footnotesize $\pm$0.5     & \uline{57.2 \footnotesize {$\pm$1.0}}  & \uline{77.2\footnotesize {$\pm$0.5}}   & \uline{80.6\footnotesize {$\pm$0.7}}  & \uline{74.3\footnotesize {$\pm$0.7}}   \\
\ourPrompt + Adv & 97.3\footnotesize $\pm$0.9     & \textbf{82.7\footnotesize $\pm$0.4}   & \textbf{90.3\footnotesize $\pm$0.2}   & \textbf{88.5 \footnotesize $\pm$1.8} & \textbf{86.4\footnotesize $\pm$0.3}   \\
\toprule
             & \multicolumn{5}{c}{MRPC}                                                  \\\toprule
             & Clean Acc      & TextFooler   & TextBugger   & BertAttack  & Misspellings       \\\midrule
\linearFT       & 87.9\footnotesize $\pm$0.6     & 41.5\footnotesize $\pm$1.2   & 50.2\footnotesize $\pm$1.0   & 61.1\footnotesize $\pm$1.1  & 51.7\footnotesize $\pm$1.0   \\
\linearFT + Adv & 87.2\footnotesize $\pm$0.4     & 42.1\footnotesize $\pm$0.3   & 53.4\footnotesize $\pm$0.7   & 64.1\footnotesize $\pm$0.1  & 54.2\footnotesize $\pm$0.4   \\
\ourPrompt       & 88.4\footnotesize $\pm$0.4     & \uline{44.8 \footnotesize {$\pm$0.1}}  & \uline{56.6\footnotesize {$\pm$0.1}}   & \uline{68.8\footnotesize {$\pm$0.5}}  & \uline{57.3\footnotesize {$\pm$0.9}}   \\
\ourPrompt + Adv & 87.1\footnotesize $\pm$1.2      & \textbf{46.6\footnotesize $\pm$1.2}   & \textbf{60.7\footnotesize $\pm$0.4}   & \textbf{72.1 \footnotesize $\pm$0.9} & \textbf{65.8 \footnotesize $\pm$0.3}  \\ \bottomrule
\end{tabular}}
\caption{Adversarial performance of RoBERTa-base model on $2$ additional datasets. All accuracy values are reported for a fixed test set of size 1000 and are averaged over $3$ different seeds. The highest accuracies are bolded, and the second-best is  underlined. 
\ourPrompt is the most robust, and preserves (or improves) the clean accuracy.}
\label{table:app:roberta-full}
\end{table*}
\begin{table*}[t]
\centering
\scalebox{\tablescale}{
\begin{tabular}{@{}lccccccc@{}}
\toprule 
& \multicolumn{7}{c}{SST2}\\ \toprule
                    & \linearFT & \linearFT + Adv & Free LB++ & MADA & InfoBert & \ourPrompt & \ourPrompt + Adv \\\midrule
Clean Acc      & 91.9 \footnotesize $\pm$0.2 & 90.9 \footnotesize $\pm$0.3 & 92.1 \footnotesize $\pm$0.8 & 92.1 \footnotesize $\pm$0.9 & 91.7 \footnotesize $\pm$0.6 & 91.7 \footnotesize $\pm$0.4 & 91.8 \footnotesize $\pm$0.7\\
TextFooler   & 38.3 \footnotesize $\pm$1.0 & 42.8 \footnotesize $\pm$1.2 & 42.2 \footnotesize $\pm$1.0 & 41.7 \footnotesize $\pm$0.5 & 43.1 \footnotesize $\pm$0.8 & \uline{44.6 \footnotesize {$\pm$0.7}} & \textbf{47.7 \footnotesize $\pm$0.6}\\
TextBugger   & 60.4 \footnotesize $\pm$0.4 & 62.3 \footnotesize $\pm$0.5 & 63.0 \footnotesize $\pm$0.7 & 60.9 \footnotesize $\pm$0.4 & 64.6 \footnotesize $\pm$0.6 & \uline{65.1 \footnotesize {$\pm$0.1}} & \textbf{67.8 \footnotesize $\pm$0.4} \\
Bertattack   & 68.7 \footnotesize $\pm$0.5 & 70.1 \footnotesize $\pm$0.8 & 72.0 \footnotesize $\pm$0.9 & 70.3 \footnotesize $\pm$0.7 & 72.8 \footnotesize $\pm$0.6 & \uline{75.9 \footnotesize {$\pm$0.7}} & \textbf{78.9 \footnotesize $\pm$0.9} \\
Misspellings & 39.2 \footnotesize $\pm$0.4 & 42.4 \footnotesize $\pm$0.4 & 43.4 \footnotesize $\pm$0.4 & 40.2 \footnotesize $\pm$0.7 & 43.1 \footnotesize $\pm$0.7 & \uline{45.6 \footnotesize {$\pm$1.1}} & \textbf{49.2 \footnotesize $\pm$0.9}\\
\toprule
& \multicolumn{7}{c}{AG News}                                               \\\toprule
& \linearFT & \linearFT + Adv & Free LB++ & MADA & InfoBert & \ourPrompt & \ourPrompt + Adv \\ \midrule
Clean Acc   & 93.7 \footnotesize $\pm$0.4 & 93.2 \footnotesize $\pm$0.2 & 93.4 \footnotesize $\pm$0.2 & 92.8 \footnotesize $\pm$0.5 & 93.8 \footnotesize $\pm$0.3 & 93.7 \footnotesize $\pm$0.5 & 94.0 \footnotesize $\pm$0.6 \\
TextFooler   & 37.5 \footnotesize $\pm$0.7 & 44.3 \footnotesize $\pm$1.0 & 43.5 \footnotesize $\pm$0.2 & 41.8 \footnotesize $\pm$0.9 & 44.0 \footnotesize $\pm$1.6 & \uline{46.3 \footnotesize {$\pm$1.2}} & \textbf{53.7 \footnotesize $\pm$0.1} \\
TextBugger   & 58.9 \footnotesize $\pm$0.6 & 64.1 \footnotesize $\pm$0.2 & 63.4 \footnotesize $\pm$0.8 & 62.6 \footnotesize $\pm$1.0 & 64.1 \footnotesize $\pm$0.8 & \uline{66.0 \footnotesize {$\pm$0.4}} & \textbf{69.2 \footnotesize $\pm$1.3} \\
Bertattack   & 78.1 \footnotesize $\pm$1.2 & 80.1 \footnotesize $\pm$0.2 & 80.9 \footnotesize $\pm$0.1 & 79.6 \footnotesize $\pm$0.6 & 80.7 \footnotesize $\pm$0.6 & \uline{82.1 \footnotesize {$\pm$0.7}} & \textbf{83.4 \footnotesize $\pm$0.4} \\
Misspellings & 76.8 \footnotesize $\pm$0.8 & 78.5 \footnotesize $\pm$0.2 & 79.5 \footnotesize $\pm$0.7 & 76.9 \footnotesize $\pm$1.3 & 79.6 \footnotesize $\pm$0.7 & \uline{81.5 \footnotesize {$\pm$0.4}} & \textbf{84.3 \footnotesize $\pm$0.3} \\
\toprule
& \multicolumn{7}{c}{BoolQ}                                               \\\toprule
& \linearFT & \linearFT + Adv & Free LB++ & MADA & InfoBert & \ourPrompt & \ourPrompt + Adv \\ \midrule
Clean Acc     & 71.1 \footnotesize $\pm$1.3 & 71.0 \footnotesize $\pm$0.9 & 70.7 \footnotesize $\pm$0.2 & 71.1 \footnotesize $\pm$0.9 & 71.8 \footnotesize $\pm$0.6 & 71.4 \footnotesize $\pm$1.0 & 71.3 \footnotesize $\pm$0.3\\
TextFooler   & 21.8 \footnotesize $\pm$4.4 & 29.8 \footnotesize $\pm$0.8 & 29.5 \footnotesize $\pm$0.6 & 25.4 \footnotesize $\pm$0.8 & 29.9 \footnotesize $\pm$0.2 & \uline{31.1 \footnotesize {$\pm$1.3}} & \textbf{43.1 \footnotesize $\pm$0.7} \\
TextBugger   & 36.8 \footnotesize $\pm$3.0 & 42.8 \footnotesize $\pm$1.3 & 42.8 \footnotesize $\pm$0.6 & 41.6 \footnotesize $\pm$0.6 & 42.6 \footnotesize $\pm$0.6 & \uline{44.4 \footnotesize {$\pm$2.8}} & \textbf{49.9 \footnotesize $\pm$0.9} \\
Bertattack   & 55.7 \footnotesize $\pm$1.2 & 57.8 \footnotesize $\pm$0.7 & 58.2 \footnotesize $\pm$0.9 & 57.8 \footnotesize $\pm$0.6 & 58.9 \footnotesize $\pm$0.8 & \uline{60.1 \footnotesize {$\pm$0.6}} & \textbf{63.2 \footnotesize $\pm$0.7} \\
Misspellings & 55.1 \footnotesize $\pm$1.0 & 58.1 \footnotesize $\pm$0.3 & 59.4 \footnotesize $\pm$0.7 & 56.2 \footnotesize $\pm$0.7 & 59.1 \footnotesize $\pm$0.6 & \uline{60.1 \footnotesize {$\pm$1.0}} & \textbf{63.2 \footnotesize $\pm$0.8} \\
\bottomrule
\end{tabular}
}
\caption{Adversarial performance of BERT-base model on $3$ different datasets. All accuracy values are reported for a fixed test set of size 1000 and are averaged over $3$ different seeds. The highest accuracies are bolded, and the second-best are  underlined. 
\ourPrompt is the most robust, and preserves (or improves) the clean accuracy.}
\label{table:app:bert-full}
\end{table*}

\paragraph{Results on BERT-Base}
Results on BERT-Base model are presented in Table~\ref{table:app:bert-full}. Similar to the results corresponding to RoBERTa-Base model in the main paper, we find that our proposed method \ourPrompt improves over the state-of-art defenses across 3 different datasets and 4 different attacks by $2\%$ even without any adversarial training. Using adversarial training further improves the average robust accuracy by $4\%$.


\paragraph{Additional Datasets}
We further extend our results on two diverse datasets---DBPedia14~\cite{dbpedia}, a 14-class news classification dataset, and MRPC~\cite{dolan-brockett-2005-automatically}, a paraphrase detection dataset. Results on these are presented for the \linearFT and \ourPrompt training schemes for RoBERTa-base model in Table~\ref{table:app:roberta-full}.

The experiments provide additional evidence to support our findings about the adversarial robustness conferred by model-tuning via prompts (\ourPromptns) as opposed to the conventional approach of \linearFTns. Without adversarial training, \ourPrompt improves over \linearFT by an average of 6\% on the MRPC dataset across 4 different attacks. 
Results on the DBPedia dataset also show consistent improvements of \ourPrompt over \linearFT. In particular, we find that \ourPrompt improves on average (across 4 different attacks) by 10\% over \linearFTns, and \ourPrompt + adv improves by 16\% over the adversarial training counterpart of \linearFTns. In a setting where the number of labels is many, we in fact see a larger relative gain by using \ourPrompt over the conventional approach of \linearFTns.

\subsection{Results on Causal Language Models}
\label{app:subsec:gpt-results}

Causal Language Models have not been traditionally fine-tuned for downstream classification tasks. This is evident also from the exclusion of fine-tuning results in the original GPT-2 paper~\citep{radford2019language}. In this work, we try to evaluate the clean and adversarial robustness of GPT-2 models, when adapted to downstream tasks. To implement \ourPromptns, we use the Causal Language Modeling (CLM) head to get the next word prediction logits. Since we are using the CLM head, it is imperative that the prompt templates are appended at the back and have the \texttt{[MASK]} as the last token.

We find that on the BoolQ dataset \linearFT achieves a robust accuracy of $20.2\%$ and \ourPrompt achieves a robust accuracy of $28.7\%$ (Table~\ref*{tab:gpt-acc}), which is a large improvement. Similar to our findings in the main paper,  $1$-step adversarial training on \ourPrompt (\ourPrompt + Adv)  yields a robust accuracy of $30.1\%$ which is a massive improvement over \linearFT and \linearFT + Adv which obtains a robust accuracy of $22.0\%$. Interestingly, we also notice that for \linearFT and \linearFT + Adv, it is difficult to achieve a good clean generalization performance whereas \ourPrompt and \ourPrompt + Adv perform much better on the clean test set. These observations are in line with the results in our main paper. On the AG News dataset, \ourPrompt performs significantly better than \linearFT and \ourPrompt + Adv performs better than \linearFT + Adv. These results show that \ourPrompt is not only a good way of finetuning BERT-like MLMs but can also improve Causal Language Models both in terms of clean accuracy and robustness to adversarial perturbations.



\subsection{Sample Efficiency and Effective Robustness}
\label{app:subsec:addn_eff}


\begin{figure}
\centering
\begin{subfigure}[t]{0.4\textwidth}
\includegraphics[width=\linewidth]{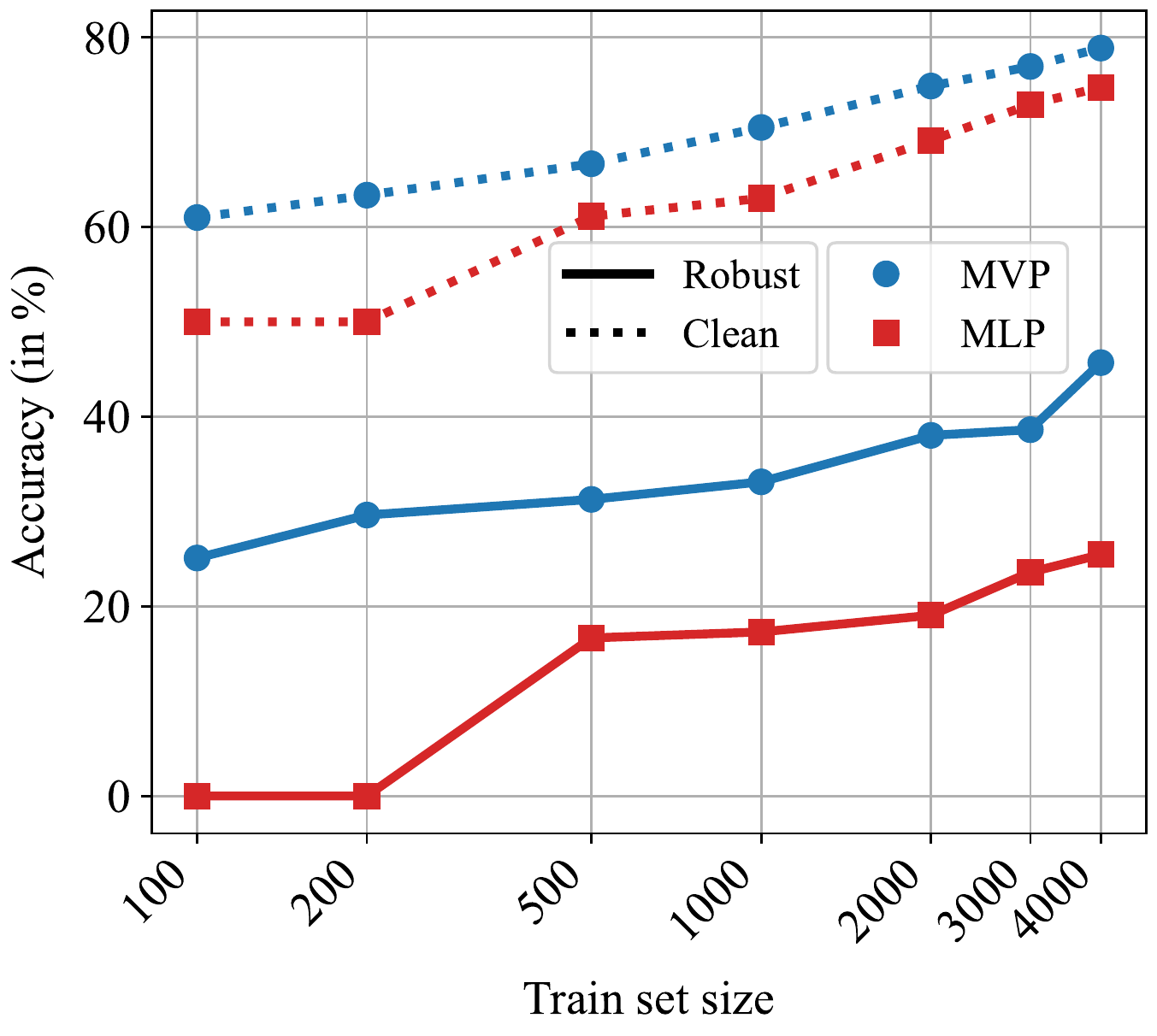}
  \caption{Clean and adversarial accuracies of RoBERTa-base model on BoolQ dataset for varying amounts of training data.}
  \label{app:fig:eff_boolq}
\end{subfigure}
\hspace{0.1\textwidth}
\begin{subfigure}[t]{0.4\textwidth}
\includegraphics[width=\linewidth]{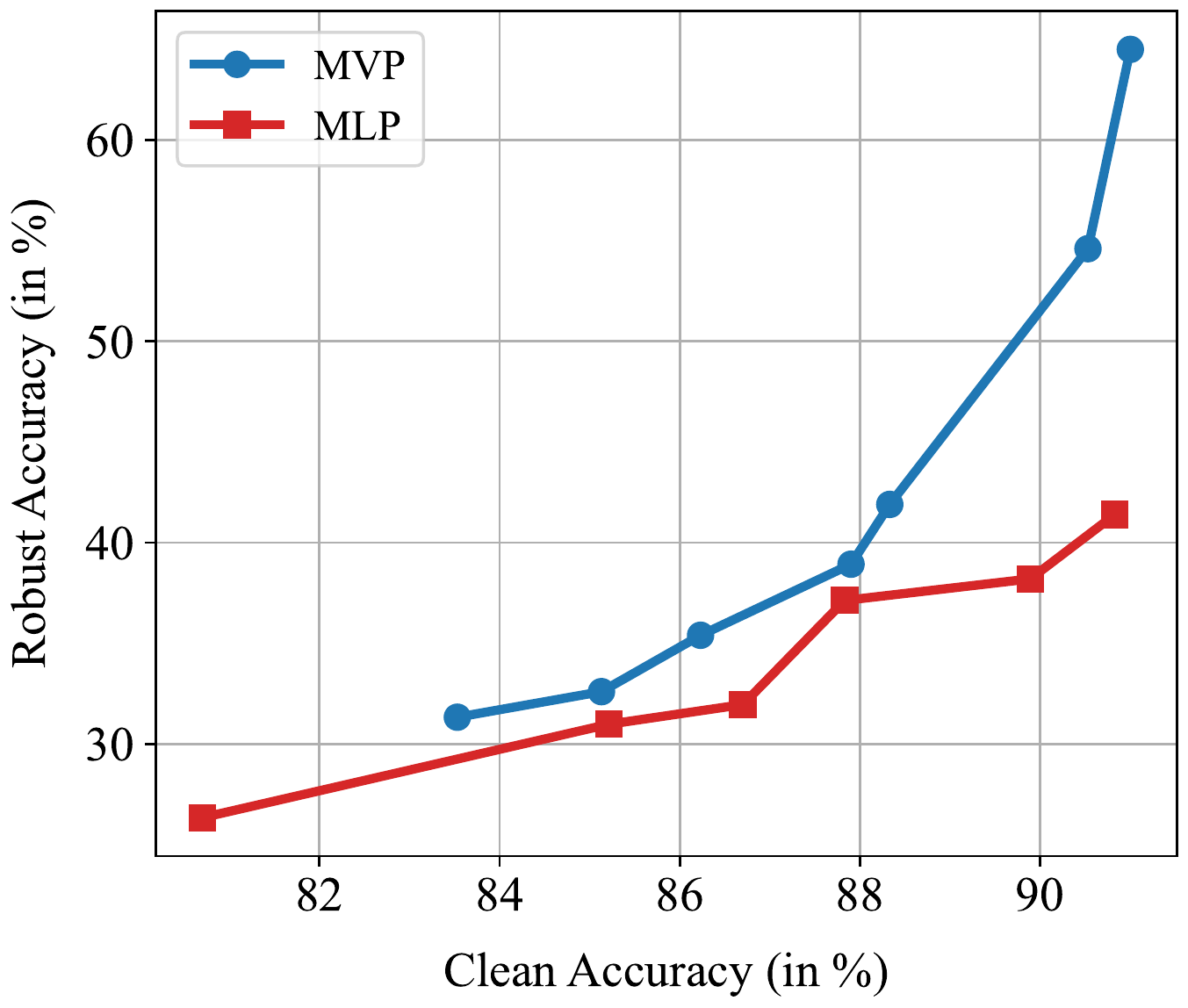}
  \caption{Clean vs adversarial performance of RoBERTa base model for the AG News dataset. We find that models tuned via prompts (\ourPromptns) yield more robust models compared to fine-tuning MLP heads for the same clean accuracy. 
  }
  \label{app:fig:scaling-curves-agnews}
\end{subfigure}
\caption{(a) Models trained with \ourPrompt are significantly more sample efficient as compared to those with \linearFT. (b) We find that models tuned via prompts (\ourPromptns) yield more robust models compared to fine-tuning MLP heads for the same clean accuracy (details in  \S\ref{subsec:additional-advantages}).}
\end{figure}

We demonstrate the sample efficiency of \ourPrompt on the BoolQ dataset (Figure \ref{app:fig:eff_boolq}) in addition to the discussion about AG News in \S\ref{subsec:additional-advantages}. Interestingly we find that \linearFT is unable to achieve better accuracy compared to even random classifiers with $200$ examples but \ourPrompt performs much better in the low data regime ($<200$ examples).  We also provide more evidence on the effective robustness of \ourPrompt by presenting the effective robustness results on AG News (Figure \ref{app:fig:scaling-curves-agnews}). Even for AG News, we notice that the curve is much steeper for \ourPrompt than \linearFTns.



\section{Extended Analysis}
\subsection{Latency of Using Multiple Templates}
\label{app:latency}

We present the latency numbers and compare them with the latency of the standard \linearFT approach. 
Specifically, the time required for 2000 forward passes of data from the IMDb dataset with a batch size of 1 is represented as $T = 24.45 \pm 0.25$ seconds. The results are presented in Table~\ref{tab:latency}.

\begin{table}[t]
    \centering
    \begin{tabular}{ll}
        \toprule
        Configuration & Time taken (in sec) \\
        \midrule
        \linearFT & \(23.25 \pm 0.37\) \((0.95 \times T)\) \\
        1 Template & \(24.45 \pm 0.25\) (T) \\
        2 Templates & \(27.51 \pm 0.45\) \((1.15 \times T)\) \\
        3 Templates & \(32.81 \pm 0.30\) \((1.34 \times T)\) \\
        4 Templates & \(35.65 \pm 0.44\) \((1.45 \times T)\) \\
        \bottomrule
    \end{tabular}
    \caption{Inference latency comparison across different configurations.}
    \label{tab:latency}
\end{table}

In summary, using multiple templates makes predictions about 1.45x slower, however, this leads to improved robustness. 

\subsection{Benefits from Prompt Tuning}
\label{app:prompt-tuning}
To assess the benefit of Prompt tuning, we conducted a series of experiments. Interestingly, even an empty template with just a \texttt{[MASK]} token, which would be considered a weak prompt, showed significant performance improvements over the standard technique of \linearFTns. We present these results for 4 different prompt choices in Table~\ref{tab:prompt_tuning}. The choice of prompts used has very little effect on model robustness in the fine-tuning regime. We tabulate the robustness results corresponding to different prompts below (for the BoolQ dataset). Here the first four prompts are the prompts we used and ``Ruby emerald \texttt{[MASK]}'' is a random prompt from vocabulary words.

\begin{table*}[h]
    \centering
    \begin{tabular}{lcc}
        \toprule
        Template & Clean Accuracy & Robust Accuracy \\
        \midrule
        Answer to the question is \texttt{[MASK]} & $81.5 \pm 0.5$ & $38.5 \pm 0.7$ \\
        \texttt{[SEP]} \texttt{[MASK]} & $81.7 \pm 0.6$ & $36.1 \pm 0.4$ \\
        I think \texttt{[MASK]} & $81.9 \pm 0.8$ & $35.9 \pm 0.2$ \\
        \texttt{[SEP]} The answer is \texttt{[MASK]} & $82.0 \pm 0.3$ & $38.1 \pm 0.1$ \\
         \texttt{[SEP]} Ruby emerald \texttt{[MASK]} & 81.4 $\pm$ 0.5 & 36.8 $\pm$ 0.4 \\
        None (\linearFTns) & $80.6 \pm 1.5$ & $28.2 \pm 1.7$ \\
        \bottomrule
    \end{tabular}
    \caption{Model robustness per template chosen for the BoolQ dataset.}
    \label{tab:prompt_tuning}
\end{table*}

We did not perform any dedicated prompt tuning for selecting the prompts. Instead, prompts were either chosen directly or inspired by the \textit{OpenPrompt} repository. The selected prompts led to a marginal (2\%) increase in model robustness during fine-tuning. Unlike typical few-shot or in-context learning methods, our approach aligns more closely with the idea of prompt tuning. For more advanced techniques and further potential improvements in prompt tuning, readers are referred to \citet{hu-etal-2022-knowledgeable}.

\subsection{Why does using "dummy candidate words" not hurt model robustness?}
\label{app:dummy-candidates}

In our paper, we note that using dummy candidate words like Jack and Ann, instead of class labels,
does not hurt model robustness. However, this is very similar to random projection layers, so why does this not impact model robustness similarly?
We note that using dummy candidate words leads to modifying an embedding of size 768 x \(C\) (where \(C\) is the number of candidate words) so that they now have a new ``meaning''. The effective number of ``new parameters'' is much lower than the parameters in the ``dense 768x768 layer'' in the Roberta model. However, in terms of new parameter complexity, this is similar to our ablation ``\sparseCLSns''. As one may note, using \sparseCLS also improves robustness over \linearFTns. This is because we avoid the dense 768x768 layer.

Additionally, we conducted a new experiment of using empty slots in the vocabulary of Roberta and compared it with using ``dummy candidate words'' and ``class labels''. For the BoolQ dataset, using a Roberta model, we summarize the results in Table~\ref{tab:dummy_words}.

\begin{table*}[h]
    \centering
    \begin{tabular}{llll}
        \toprule
        Method & Candidate Choice & Clean Accuracy & Robust Accuracy \\
        \midrule
        \linearFT & N/A & 80.6 \(\pm\) 1.5 & 28.2 \(\pm\) 1.6 \\
        ProjectCLS & N/A & 81.3 \(\pm\) 0.5 & 37.4 \(\pm\) 1.2 \\
        \ourPrompt & Class Labels & 82.0 \(\pm\) 0.6 & 42.9 \(\pm\) 0.5 \\
        \ourPrompt & Dummy Candidate Words (Jack/Ann) & 80.9 \(\pm\) 0.3 & 42.1 \(\pm\) 0.4 \\
        \ourPrompt & Empty Slots & 81.3 \(\pm\) 0.3 & 41.2 \(\pm\) 0.7 \\
        \bottomrule
    \end{tabular}
    \caption{Comparison of different choices of candidate words and their accuracies when training a Roberta model on the BoolQ dataset.}
    \label{tab:dummy_words}
\end{table*}

Based on these accuracies above we find that:

\begin{enumerate}
    \item Using class labels is better than using “dummy/untrained words” for both clean and robust accuracy, which supports the random parameter vulnerability hypothesis.
    \item The robustness achieved upon using completely untrained slots is similar to that when using dummy candidate words. This suggests that when compared to class labels, modifying dummy words has a similar loss in robustness as with modifying untrained words.
    \item The \ourPrompt models (with random/empty candidate words) are more robust than \sparseCLS (which already bridges the robustness gap from \linearFTns). These gains are explained by the pre-training task alignment hypothesis, where pre-training (and fine-tuning) the model with the task of \texttt{[MASK]} infilling helps make the downstream model robust.
\end{enumerate}

\subsection{Impact of Ensembling the Candidates}
\label{app:enemble-candidates}

Recall that in the main paper, we ensemble multiple templates and aggregate their predictions. In this subsection, we also investigate the impact of ensembling candidate words rather than templates. Based on the results in Table~\ref{tab:ensemble-candidates}, we find that this is not as helpful as ensembling multiple templates.

\begin{table*}[h]
\centering
\begin{tabular}{@{}lcc@{}}
    \toprule
    {Configuration} & {Clean Accuracy} & {Robust Accuracy} \\
    \midrule
    1 prompt + 4 candidates & $81.9 \pm 0.3$ & $37.4 \pm 0.5$ \\
    1 prompt + 1 candidate & $81.9 \pm 0.8$ & $35.9 \pm 0.2$ \\
    4 prompt + 1 candidate & $82.0 \pm 0.6$ & $42.9 \pm 0.5$ \\
    \bottomrule
\end{tabular}
\caption{Impact of different ensembling configurations on clean and robust accuracy of Roberta model on the BoolQ dataset.}
\label{tab:ensemble-candidates}
\end{table*}

\section{Hyperparameter Details}
\label{appendix:hyperparameters}
\paragraph{Attack Hyperparameters}
TextFooler and TextBugger use a word substitution attack that searches for viable substitutions of a word from a set of synonyms. We restrict the size of the synonym set to $50$ for TextFooler which is the default value used by \citet{Jin_Jin_Zhou_Szolovits_2020} and to $5$ which is the default value used by \citet{li2018textbugger}. Both TextFooler and TextBugger use a Universal Sentence Encoder (USE), that poses a semantic similarity constraint on the perturbed sentence. We use the default value of $0.84$ as the minimum semantic similarity. Another important attack parameter is the maximum percentage of modified words ($\rho_{\text{max}}$). As discussed in \citep{li-etal-2021-searching}, we use $\rho_{\text{max}} = 0.3$ for AG News and use $\rho_{\text{max}} = 0.1$ for BoolQ and SST2 in all our experiments. We use a query budget of $100$ for BERT-Attack and a query budget of $300$ for adversarial misspellings as these attacks are very slow.

\paragraph{Training Hyperparameters \& Model Selection}

We train all models including the baselines with patience of $10$ epochs, for a maximum of $20$ epochs, and choose the best model based on validation accuracy. For the datasets that do not contain a publicly available validation set, we set aside $10\%$ of the training set for validation. In the case of baseline defenses that use adversarial training, we perform model selection based on adversarial accuracy rather than clean accuracy. We use a candidate answer set containing only the class label names and we average over $4$ prompt templates in all the \ourPrompt models. We use a batch size of $32$ for \linearFT and a batch size of $8$ for \ourPrompt models. The learning rate is set as $1$e$-5$ for all the models. We use the AdamW optimizer along with the default linear scheduler~\citep{wolf-etal-2020-transformers}. In all the \ourPrompt + Adv and \linearFT + Adv models, we use a use $1$-step adversarial training with max $\ell_2$ norm of $1.0$. For the state-of-the-art baselines, we use the same hyperparameters as prescribed by the original papers.

\section{Human Study}
\label{sec:discussion}
Despite the improvements brought to adversarial robustness by our proposed modification (\ourPrompt + Adv), we note that there is still a significant drop in robust accuracy as opposed to the clean accuracy of the model. 
We conduct a human study in order to (i) assess the viability of adversarial attacks, and (ii) estimate human performance against adversarial attacks. More specifically, we provide machine learning graduate students $250$ input examples and ask the following questions: 

\begin{enumerate}
    \item What is the perceived label of the sentence? (Answer options: True or False)
    \item On a scale of $1$ to $3$, what is their confidence about this label? 
    \item Was this sentence adversarially manipulated? 
    (Answer options: Yes, Unsure, or No)
\end{enumerate}
\begin{table}[t]
\centering
\scalebox{\tablescale}{
\begin{tabular}{@{}llcc@{}}
\toprule
\multicolumn{2}{l}{}                                  & \multicolumn{1}{l}{\multirow{2}{*}{\begin{tabular}[c]{@{}c@{}}Perturbed \\ Examples\end{tabular}}} & \multicolumn{1}{l}{\multirow{2}{*}{\begin{tabular}[c]{@{}c@{}}Unperturbed\\ Examples\end{tabular}}} \\
                               &                      & \multicolumn{1}{l}{}                                                                                 & \multicolumn{1}{l}{}                                                                             \\ \midrule
\multicolumn{2}{l}{Q1. Annotator Accuracy}   & 74\%                            & 85\%                         \\
\multicolumn{2}{l}{Q2. Annotator Confidence} & 75\%                            & 90\%                         \\
\cmidrule{1-4}
\multicolumn{1}{l}{
\multirow{3}{*}{Q3. Perturbed?}}   & No       & 54\%                            & 82\%                         \\
                                 & Unsure   & 17\%                            & 12\%                         \\
                                 & Yes      & 29\%                            & 06\%                          \\ \bottomrule
\end{tabular}}
\caption{Summary of the responses from the user study. The total number of presented examples is 250, out of which 83 are unperturbed and 167 are adversarially perturbed.}
\label{tab:human-study-main}
\end{table}
We use the BoolQ dataset and strictly instruct our annotators 
to not use any external knowledge but the only context of the given passage. We use samples that were successfully attacked by TextFooler for \ourPrompt + Adv model with a RoBERTa backbone. As a control for the study,  33\% of all sentences are unperturbed sentences from the original dataset. The underlying model achieves a clean accuracy of $81.7\%$ and a robust accuracy of $54.0\%$. 


First, we find that humans achieved $11\%$ lower accuracy
on adversarial examples as compared to clean examples
($85\% \to 74\%$) with average confidence on the label
of perturbed examples being $15\%$ lower ($90\% \to 75\%$) (Table \ref{tab:human-study-main}).
Next,
we also discover that human annotators suspect $29\%$ of adversarial examples to be perturbed as opposed to only $6\%$ of clean examples. 
Through this study, we also find 
that in $47\%$ of the cases,
the input is either manipulated so significantly that it is easily detectable or the original label is not preserved, signifying that \ourPrompt may be more robust than what statistics suggest in \S\ref{sec:all-results}. Further details are available in Appendix \ref{app:sec:human-study}.

\subsection{Details of Interface}
\label{app:sec:human-study}

\begin{figure}[t]
\centering
\includegraphics[width=1\linewidth]{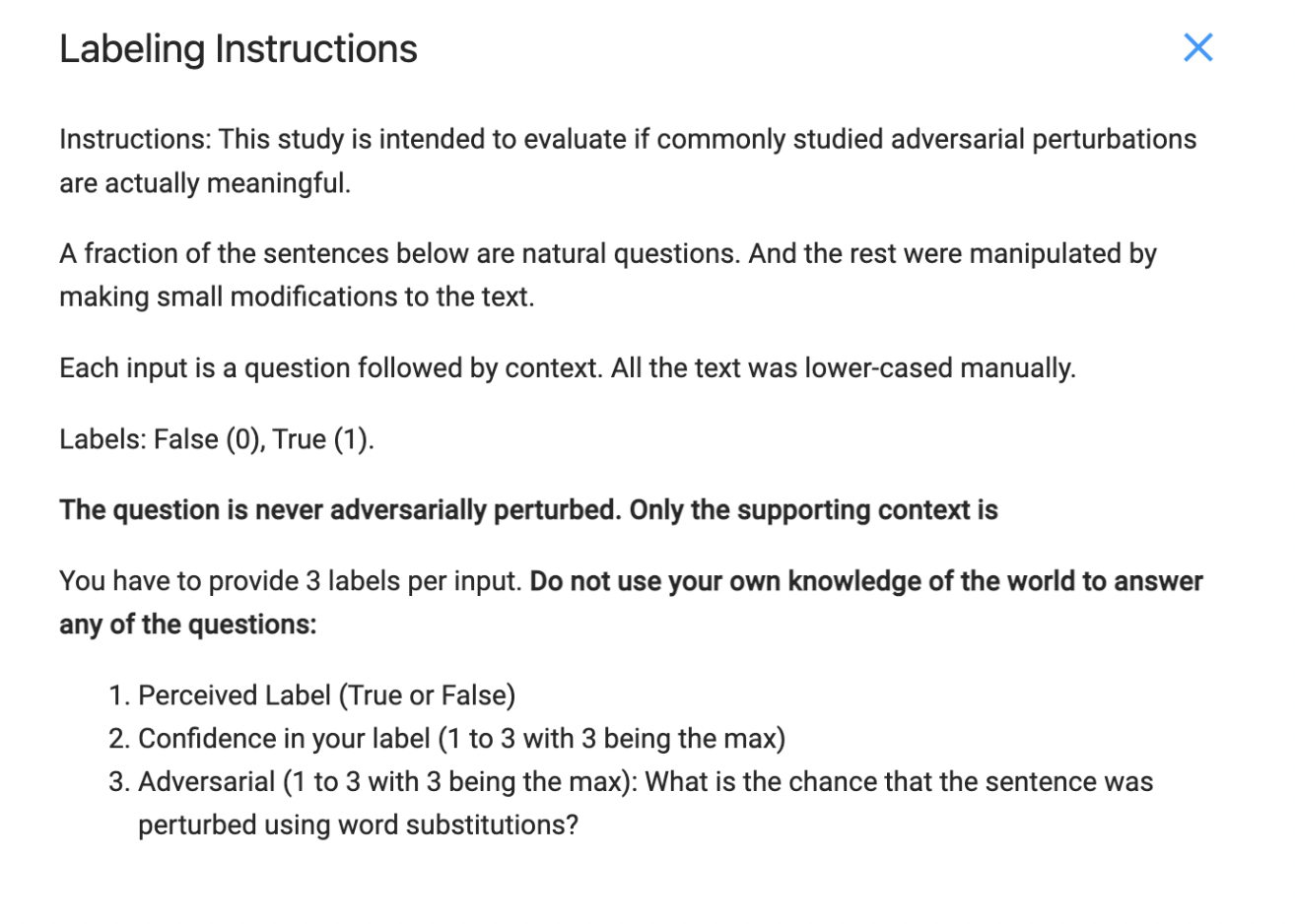}
  \caption{A snapshot of the instructions for completing our study.}
  \label{app:fig:instructions}
\end{figure}


We present a snapshot of our interface that provides detailed instructions for our users (Figure ~\ref{app:fig:instructions}).  We provide a detailed overview of the questions asked in the user study. Annotators were provided with a boolean question and an accompanying context to answer the question and asked were asked to annotate the following:
\paragraph{1. What should be the answer to the question? (only use the context)} Given the boolean question and the context, we ask the annotators whether the answer to the question is True or False. We also request the annotators only use the given context and refrain from using any external knowledge.

\paragraph{2. How confident are you about the label above?} Once the annotator has answered question 1, we ask them to rate how confident they feel about the label they assigned to the input. The options provided are "Uncertain", "Somewhat Certain" and "Certain". Based on their response we assign a confidence of $1$, if the annotator was certain, assign $0.5$ if the annotator was somewhat certain, and assign $0$ if the annotator was uncertain to calculate the average confidence.

\paragraph{3. Do you think that the sentence is adversarially perturbed? (using word substitutions) Do not use your own knowledge of the world to answer this question.}We also ask the annotators, if the input was adversarially perturbed. The options provided to the user are "No", "Unsure" and "Yes". 

The annotators helped us annotate $250$ such examples out of which $167$ were adversarially perturbed and $83$ were clean. An overview of the responses from this study is presented in Table~\ref{tab:human-study-main}, and the key takeaways are discussed in Section~\ref{sec:discussion}. 



\end{document}